\pdfoutput=1

\documentclass[11pt]{article}

\usepackage{EMNLP2022}
\usepackage{times}
\usepackage{latexsym}
\usepackage{graphicx}
\usepackage{booktabs}
\usepackage{multirow}
\usepackage{fontawesome}
\usepackage{soul}
\usepackage{longtable}
\usepackage{xcolor}

\usepackage[utf8]{inputenc}

\usepackage{microtype}

\usepackage{inconsolata}

\usepackage{color, colortbl}
\usepackage{xspace}
\usepackage{float}
\usepackage{amsmath}
\usepackage{algorithmicx}
\usepackage{algorithm}
\usepackage{algpseudocode}
	
\definecolor{highlightclr}{rgb}{0.95,0.95,0.95}
\definecolor{forestgreen}{rgb}{0.0,0.37,0.0}
\definecolor{lightred}{rgb}{0.5,0.0745,0.192}

\newcommand{\ph}[1]{#1}
\newcommand{\pr}[1]{\textcolor{lightred}{#1}}
\newcommand{\pg}[1]{\underline{\textcolor{forestgreen}{#1}}}

\newcommand{\pgb}[1]{\underline{\textbf{\textcolor{forestgreen}{#1}}}}
\newcommand{\pt}[1]{\phantom{#1}}
\newcommand{\bx}[1]{\textbf{#1}} 

\newcommand*{\img}[1]{%
    \raisebox{-.1\baselineskip}{%
        \includegraphics[
        height=0.85\baselineskip,
        width=0.85\baselineskip,
        keepaspectratio,
        ]{#1}%
    }%
}
 
\newcommand{\teabreac}{\textsc{TeaBReaC}\xspace}
\newcommand{\icon}{\img{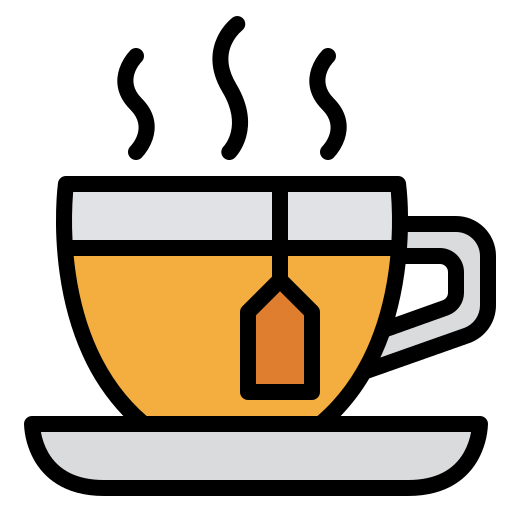}}
\newcommand{\eat}[1]{}
\newcommand{\ateforemnlp}[1]{}
\newcommand{\camerareadynote}[1]{}

\newcommand{\preasm}{PReasM\xspace}
\newcommand{\poet}{POET\xspace}
\newcommand{\numglue}{NumGLUE\xspace}
\newcommand{\ntfive}{NT5\xspace}
\newcommand{\iircg}{IIRC-G\xspace}
\newcommand{\iircr}{IIRC-R\xspace}

\usepackage{pythonhighlight}

\title{\icon \xspace Teaching Broad Reasoning Skills for Multi-Step QA \\ by Generating Hard Contexts}

\newcommand{\affa}{$^\dagger$}
\newcommand{\affb}{$^\ddagger$}

\author{
  Harsh Trivedi\affa\thanks{\enskip The work was done during the first author’s internship at the Allen Institute for AI.} \ \ \ Niranjan Balasubramanian\affb \ \ \\
  \\
  \affa Stony Brook University\\
  Stony Brook, U.S.A.\\
  \texttt{\small \{hjtrivedi,niranjan\}@cs.stonybrook.edu}
  \And
  \hspace{5ex} Tushar Khot\affb \ \ \ Ashish Sabharwal\affb \\
  \\
  \hspace{5ex} \affb Allen Institute for AI\\
  \hspace{5ex} Seattle, U.S.A.\\
  \hspace{5ex} \texttt{\small \{tushark,ashishs\}@allenai.org}
}

\begin{document}

\maketitle

\begin{abstract}
Question-answering datasets require a broad set of reasoning skills. We show how to use question decompositions to teach language models these broad reasoning skills in a robust fashion. Specifically, we use widely available QDMR representations to programmatically create hard-to-cheat synthetic contexts for real questions in six multi-step reasoning datasets. These contexts are carefully designed to avoid reasoning shortcuts prevalent in real contexts that prevent models from learning the right skills. This results in a pretraining dataset, named TeaBReaC, containing 525K multi-step questions (with associated formal programs) covering about 900 reasoning patterns. We show that pretraining standard language models (LMs) on TeaBReaC before fine-tuning them on target datasets improves their performance by up to 13 F1 points across 4 multi-step QA datasets, with up to 21 point gain on more complex questions. The resulting models also demonstrate higher robustness, with a 5-8 F1 point improvement on two contrast sets. Furthermore, TeaBReaC pretraining substantially improves model performance and robustness even when starting with numerate LMs pretrained using recent methods (e.g., PReasM, POET). Our work thus shows how to effectively use decomposition-guided contexts to robustly teach multi-step reasoning.\footnote{Code and data available at \url{https://github.com/stonybrooknlp/teabreac}.}
\end{abstract}
\section{Introduction}

Multi-step Question Answering (QA) is a complex problem that requires a wide variety of reasoning skills. In addition to basic reading comprehension (RC), models must connect multiple pieces of information, sometimes employ numerical and other forms of discrete reasoning, and compose these skills as needed for the question. However, even though questions in multi-step datasets often cover a broad range of interesting reasoning patterns, most questions follow only a few patterns, which is what models trained on these datasets naturally focus on. Moreover, the contexts occurring in existing RC datasets often contain artifacts and reasoning shortcuts~\cite{min2019compositional,chen2019understanding,dire}. Such contexts allow models to find the answer while bypassing some reasoning steps, in turn preventing models from learning the intended reasoning skills.

\begin{figure}
    \advance\leftskip-0.4em
    \centering
	\includegraphics[width=0.48\textwidth]{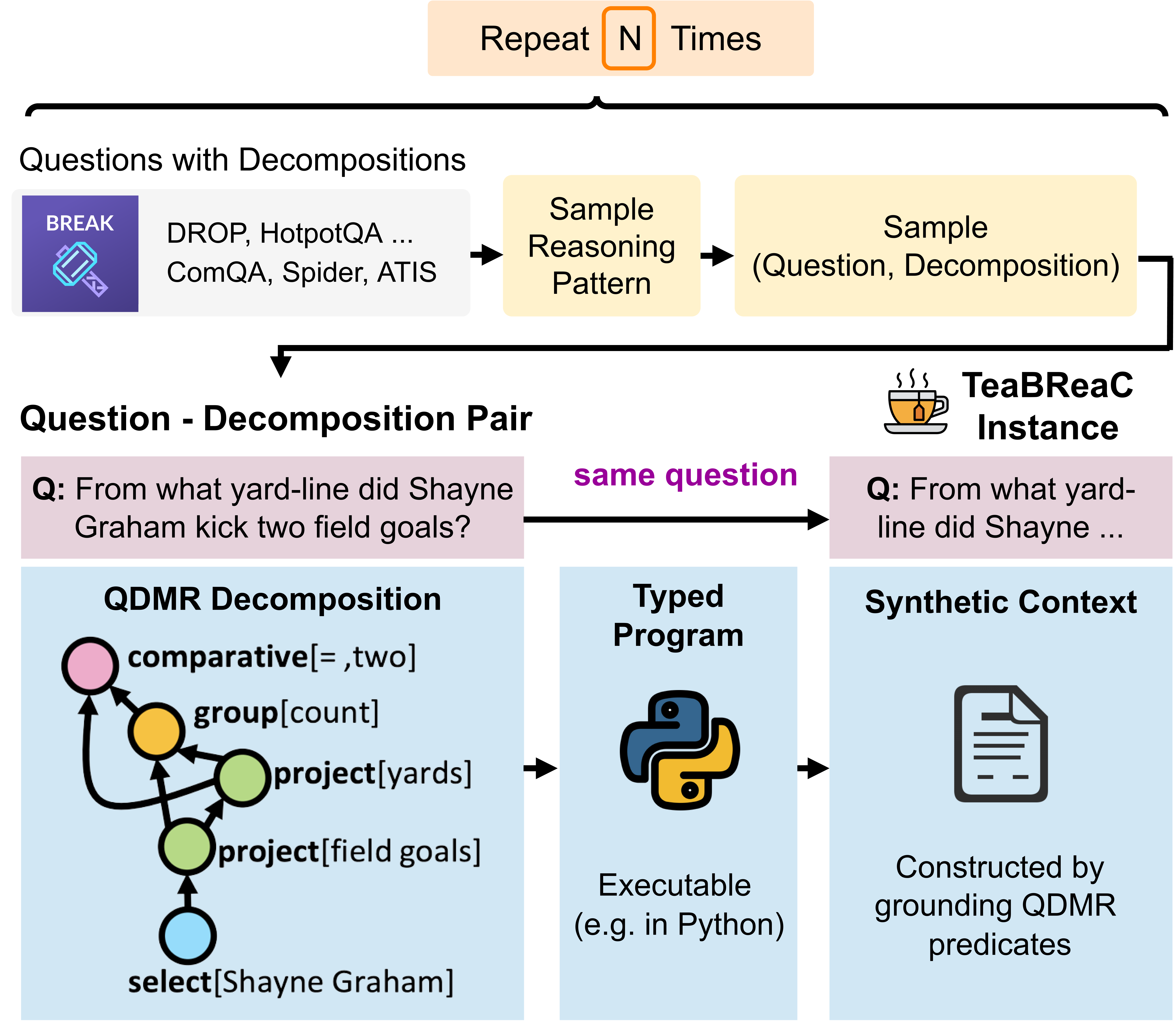}
	\caption{\teabreac \icon\ Dataset Construction: We use widely available question decomposition annotations (QDMRs) for real questions from a broad range of datasets to carefully construct synthetic contexts such that answering the resulting \icon\ question requires proper multi-step reasoning. These questions are further re-balanced to help teach a broad set of reasoning skills.}
	\label{fig:introduction}
	\vspace{-2ex}
\end{figure}

How, then, can we teach models broad multi-step reasoning skills? One way is to have greater control over the distribution of reasoning patterns and the types of input contexts models see during training---contexts that don't allow models to easily succeed via shortcuts. We observe that questions in existing datasets (henceforth referred to as ``real questions'') already cover a wide variety of reasoning patterns. The challenge, then, is to teach these reasoning patterns \emph{robustly}, even when they are relatively rare (e.g., 4-6 step reasoning). As a means to this end, we turn to \emph{synthetic context generation for real questions}. Specifically, we propose to construct contexts for real questions synthetically from scratch (instead of perturbing existing contexts), resulting in much greater control over reasoning shortcuts. Further, context generation also enables us to balance out the distribution of reasoning patterns, e.g., by synthesizing additional contexts (and thereby examples) for questions from the long-tail of underrepresented reasoning patterns.

Our use of synthetic contexts to reliably teach broad skills is inspired by three strands of recent RC QA research. One strand has shown that skills learnt over synthetic data can indeed transfer to real datasets~\cite{genbert,nt5,preasm,poet}. A second strand has shown that perturbing the existing (natural) contexts of RC instances in a targeted fashion can reduce artifact-based reasoning~\cite{jia2017adversarial,dire}. A third strand has shown that carefully constructing contexts (for synthetic questions) to have sufficient distractors can reduce artifacts~\cite{musique,commaqa}.

Building upon these three strands, we introduce \teabreac,\footnote{\teabreac $=$ ``\textbf{T}eaching \textbf{B}road \textbf{Rea}soning skills via decomposition-guided \textbf{C}ontexts''; pronounced ``Tea Break".} a teaching dataset that includes carefully constructed synthetic contexts for a broad set of real multi-step questions sourced from six existing datasets. \teabreac was designed with the goals of strong control over cheatability and balanced coverage of reasoning patterns. To identify the intended reasoning, we leverage question decomposition annotations, specifically Question Decomposition Meaning Representation or QDMR annotations which are widely available for a broad set of datasets~\cite{break}.

Figure~\ref{fig:introduction} shows the overview of our construction process for \teabreac. Our approach relies on treating a question decomposition as an unambiguous typed program that can be used to generate a synthetic context and can be executed to provide an answer. To this end, we first turn natural language QDMRs into a precise typed \textit{program}. We then construct a synthetic context by asserting a set of facts that relate to various parts of the multi-step question. We do this by grounding the predicates of QDMR (e.g., \textit{field goals of Shayne Graham} in Fig.~\ref{fig:introduction}) with randomly generated entities. We also add distractor statements to the context to ensure that bypassing reasoning steps results in an incorrect answer. The resulting contexts are hard to cheat on and thereby force models to learn the intended reasoning. We then add an outer loop around this process that ensures that the reasoning patterns---as measured by the program signatures of the questions---remain balanced in the final dataset. This forces models to learn a broad range of reasoning patterns instead of focusing on the few dominant ones. Finally, similar to prior work~\cite{genbert}, we also add simpler single-step questions to teach individual primitive skills underlying our formal programs.

Our experiments demonstrate that pretraining\footnote{We use the word pretraining to mean fine-tuning on our generated QA data before (hence \emph{pre}training) fine-tuning on the target dataset.} large language models (LMs) on \teabreac before fine-tuning on target multi-step QA datasets results in significant improvements on multiple in-distribution evaluation sets (DROP~\cite{drop}, TAT-QA~\cite{tatqa}, IIRC~\cite{iirc}), NumGLUE~\cite{numglue} by up to 13 F1 points, as well as on two contrastive evaluation sets of DROP by 5-8 points. Furthermore, even if we start with numerate LMs already pretrained on similar past work~\cite{genbert,nt5,preasm,poet}, \teabreac provides further improvement by up to 11 F1 points. Interestingly, \teabreac is substantially more beneficial for more complex questions (those with more reasoning steps), improving the T5-Large model by about 20 F1 points on questions with 5 or more steps. More generally, we expect \teabreac to be most valuable for datasets that require complex aggregation operations and their diverse compositions.

In summary, we make three contributions:

(1) A novel methodology to create a teaching dataset (a) with broad reasoning skills covering a wide range of multi-step reasoning patterns and (b) leveraging existing QDMR annotations to carefully construct contexts that require true multi-step reasoning.
(2) The \teabreac teaching dataset with over 525K questions covering about 900 reasoning patterns or program signatures. (3) An empirical demonstration that pretraining on \teabreac before fine-tuning makes both regular and numerate LMs much more effective and robust at multi-step reasoning, especially for more complex questions.

\section{Related Work}
Question Decompositions have been used to build stronger models~\cite{complexwebqns,decomprc,tmn} and challenge evaluation sets by modifying the questions~\cite{bpb}. In contrast, our goal in this work is to use decompositions to teach broad multi-step reasoning skills to any text-to-text model by creating challenging contexts for real questions.

Building synthetic datasets to teach requisite skills has been considered in prior work, but limited to only numeric reasoning skills~\cite{genbert,nt5} or few templated multi-step reasoning patterns~\cite{preasm,pan-etal-2021-unsupervised}. Even pretraining on program executions (arithmetic, logic-based, and SQL-based) has been shown to help on multi-step QA tasks~\cite{poet}. In this work, we use real questions from a wide variety of datasets and show larger gains than these prior models. We even improve these prior models by fine-tuning on our dataset.

We create more robust models by teaching reasoning skills via a dataset carefully designed to avoid shortcuts. Past work often focuses on identifying lack of robustness via analysis~\cite{min2019compositional,dire} or challenge evaluation sets~\cite{jiang2019avoiding,bpb}.

Lastly, we define new conditions for constructing contexts for real questions with minimal reasoning shortcuts. This differs from prior work that only provides conditions to measure reasoning shortcuts in \emph{existing} datasets~\cite{dire}. The ``MuSiQue condition'' of \citet{musique} targets the construction of \emph{new} non-cheatable multi-step datasets. We enforce this condition in \teabreac and introduce two additional ones that are especially pertinent to our construction. Appendix~\ref{app:related-work} includes additional discussion.

\section{Teaching Broad-Coverage Reasoning Skills in a Robust Fashion}

Multi-step questions come in a wide variety. Some involve numeric operations~\cite{drop}, some involve assessing whether complete information is present or not~\cite{iirc}, some involve tables and text~\cite{tatqa}, and so on. One way to surface the reasoning needed for answering these questions is to look at their \emph{decomposition} into smaller reasoning steps. \ateforemnlp{that can be composed together in order to arrive at the correct answer.} E.g., consider the question in Fig.~\ref{fig:introduction}, \emph{From what yard-line did Shayne kick two field goals?}. This can be decomposed as follows: list the field goals by Shayne Graham, identify the yard-lines for each of them, map each yard-line with the field goal and count them, and select the yard-line with two field goals.

While questions in multi-step QA datasets are authored with the intent that such multi-step reasoning will be used to answer them, the context associated with the questions often allows models to cheat by taking shortcuts~\cite{min2019compositional,chen2019understanding,dire}. E.g., if the context mentions field goals only by Shayne Graham and no one else, models can ignore the player name and still succeed.

Our key observation is that the decomposition of a question can be leveraged to carefully design a synthetic context for it that is hard to cheat, thereby allowing us to teach models a broad range of reasoning skills in a robust fashion. To achieve this, we procedurally create a large pretraining RC dataset, \teabreac, by using real multi-step questions (from existing datasets) and their decompositions (available in the form of QDMRs), and carefully building synthetic contexts.

QDMR or Question Decomposition Meaning Representation~\cite{break} is a common way to represent the reasoning in many types of multi-step questions as a structured decomposition graph. QDMR has standardized operators (represented as nodes) such as \texttt{select}, \texttt{project}, \texttt{group}, etc., that transform their input. These are connected together to a final node which produces the answer. Figure~\ref{fig:introduction} shows the above example question paired with its QDMR graph. Importantly, QDMRs are already available for several multi-step QA datasets.

Briefly, our method involves the following main steps; these are described in more detail in \S\ref{sec:construction}.

\paragraph{Making QDMRs more precise.}

To create QA instances that teach the precise reasoning in QDMRs, we need a precise and formal representation of reasoning captured in QDMRs. QDMRs, although structured, don't quite do so, as they are written in natural language and don't specify input/output datatypes. Since this is crucial for our approach, we convert QDMRs into formal programs with over 44 executable primitive operations along with their input/output types (\S~\ref{subsec:instance-generator}).

\paragraph{Teaching robust compositional skills.}

Past work has shown that compositional questions don't necessitate multi-step reasoning as datasets often have reasoning shortcuts~\cite{min2019compositional,chen2019understanding,dire}. To teach the reasoning reflected our formal programs robustly, our QA instances must be such that  models cannot bypass the reasoning steps and still arrive at the correct answer. To achieve this goal, we create a synthetic QA instance from a question-program pair, where the question is the same as the original question, but the context is procedurally constructed by grounding the predicates in QDMR in a careful way such that models can't cheat their way to the correct answer.

\paragraph{Teaching a broad range of reasoning patterns}

Although QDMRs cover a broad range of reasoning patterns, we find that the natural distribution of reasoning patterns in QDMRs is extremely skewed towards popular reasoning patterns (\S~\ref{subsec:dataset-generator}). Training on QA instances generated from such a distribution leads models to overfit to only a few most representative reasoning patterns, and not learn broad-range reasoning skills. To ensure this doesn't happen, we make sure our synthetic dataset is more balanced in terms of reasoning patterns (\S~\ref{subsec:dataset-generator}).

\paragraph{Teaching a broad range of reasoning primitives.}

In addition to our process of constructing a pretraining dataset to teach compositional skills described thus far, we observe that it also helps if we teach models the constituent primitive reasoning skills. To achieve this, similar to prior work~\cite{genbert}, we procedurally generate QA instances based on fixed templates for each of the 44 primitives present in our formal programs (\S~\ref{subsec:primitive-data-generator}).
\section{\teabreac Dataset Construction}
\label{sec:construction}

\begin{figure}[t]
    \centering
	\includegraphics[width=0.47\textwidth]{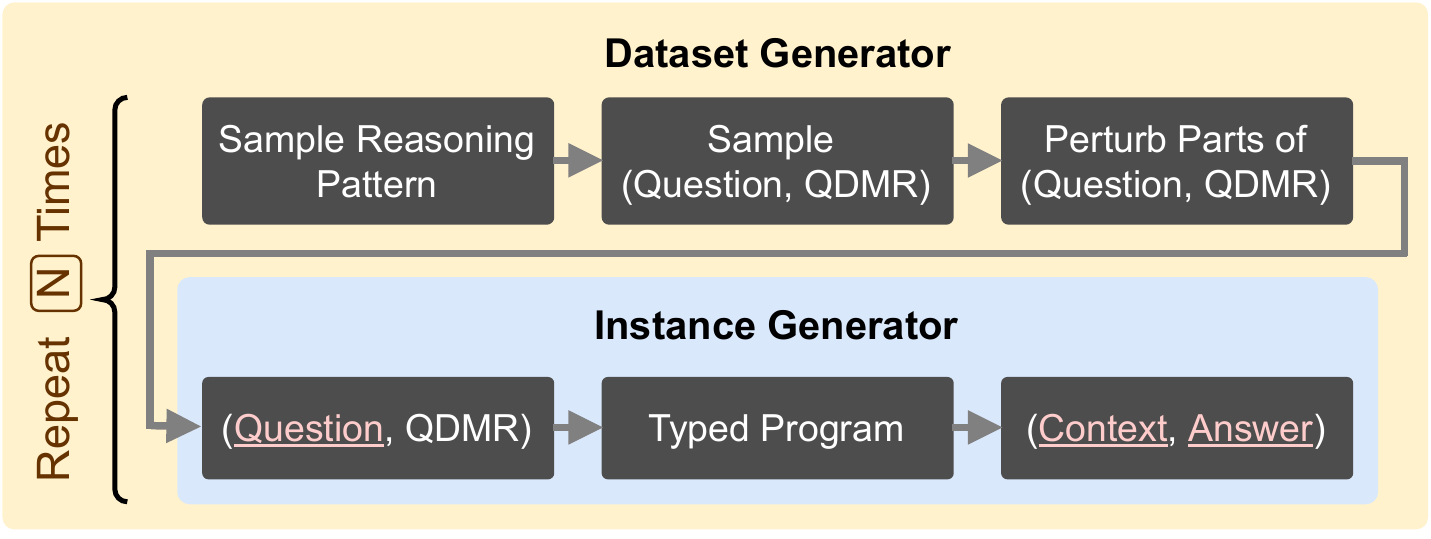}
	\caption{Schematic of \teabreac construction.}
	\label{fig:pipeline}
\end{figure}

The overview of \teabreac construction pipeline is shown in Fig.~\ref{fig:pipeline}. We discuss the QA instance generator in \S~\ref{subsec:instance-generator} and the dataset generator in \S~\ref{subsec:dataset-generator}.

\subsection{Instance Generator}
\label{subsec:instance-generator}

The Instance Generator takes a question $Q$ and its QDMR decomposition $D$ as input, and generates a synthetic context $C$ and the corresponding answer $A$ as its output. The tuple ($Q$, $C$, $A$) is the generated RC QA instance. This conversion happens in two steps: (i) QDMR to Typed Program, (ii) Typed Program to Context and Answer.

\subsubsection{QDMR to Typed Program:} 

Our goal is to generate a synthetic context $C$ that can be used to answer the question $Q$ (based on the QDMR $D$), and to also provide the answer $A$. To generate $C$ and $A$, we must be able to create facts corresponding to steps in the QDMR reasoning graph (i.e., ground the QDMR predicates\footnote{E.g., the step \texttt{``return players who kicked \#1"} has the predicate \texttt{``return players who kicked \_\_"}.}) and compute the final answer by stepping through it.

To achieve this, we need a formal representation (\textbf{Program}) that captures the precise reasoning implied by $D$, and that can be executed step-by-step (e.g., in a programming language like Python). This isn't possible directly via QDMRs as (i) although structured, they are written in natural language and have variation inherent in natural language; (ii) they don't have input and output type information, e.g., it is unclear whether the \texttt{project} operator should generate a dictionary, a list, or a scalar, making it difficult to make execute it.

\begin{figure}[ht]
    \centering
	\includegraphics[width=0.45\textwidth]{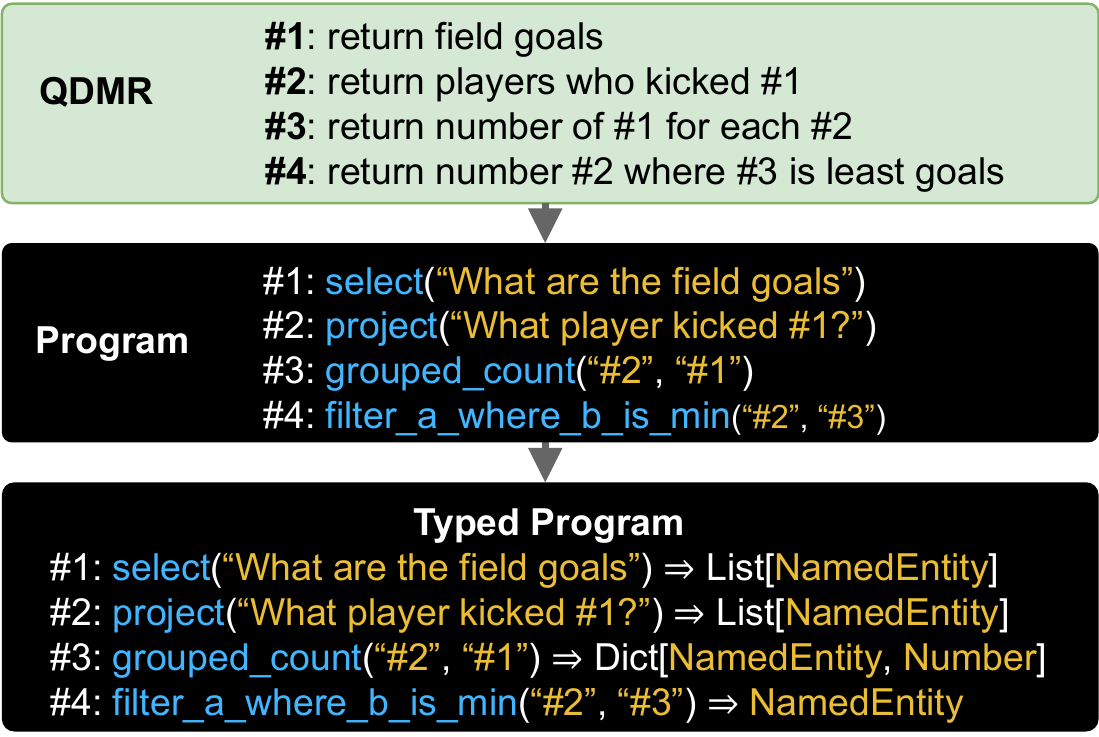}
	\caption{Example conversion of a QDMR decomposition (top) to a Typed Program (bottom).}
	\label{fig:qdmr-to-typed-program}
\end{figure}

To convert a QDMR $D$ into a Program $P$, we define a set of python functions (primitives)\footnote{We have 44 primitives operating over various types (number, date, named entity) and structures (scalar, list, dictionary) of inputs and outputs. The full list is given in App.~\ref{sec:primitives-list}.} like \texttt{select}, \texttt{filter}, \texttt{grouped\_count}, etc, and parse QDMRs into these functions using rules and heuristics. An example conversion is shown in Fig.~\ref{fig:qdmr-to-typed-program}.

\begin{figure*}[t]
    \centering
	\includegraphics[width=0.85\textwidth]{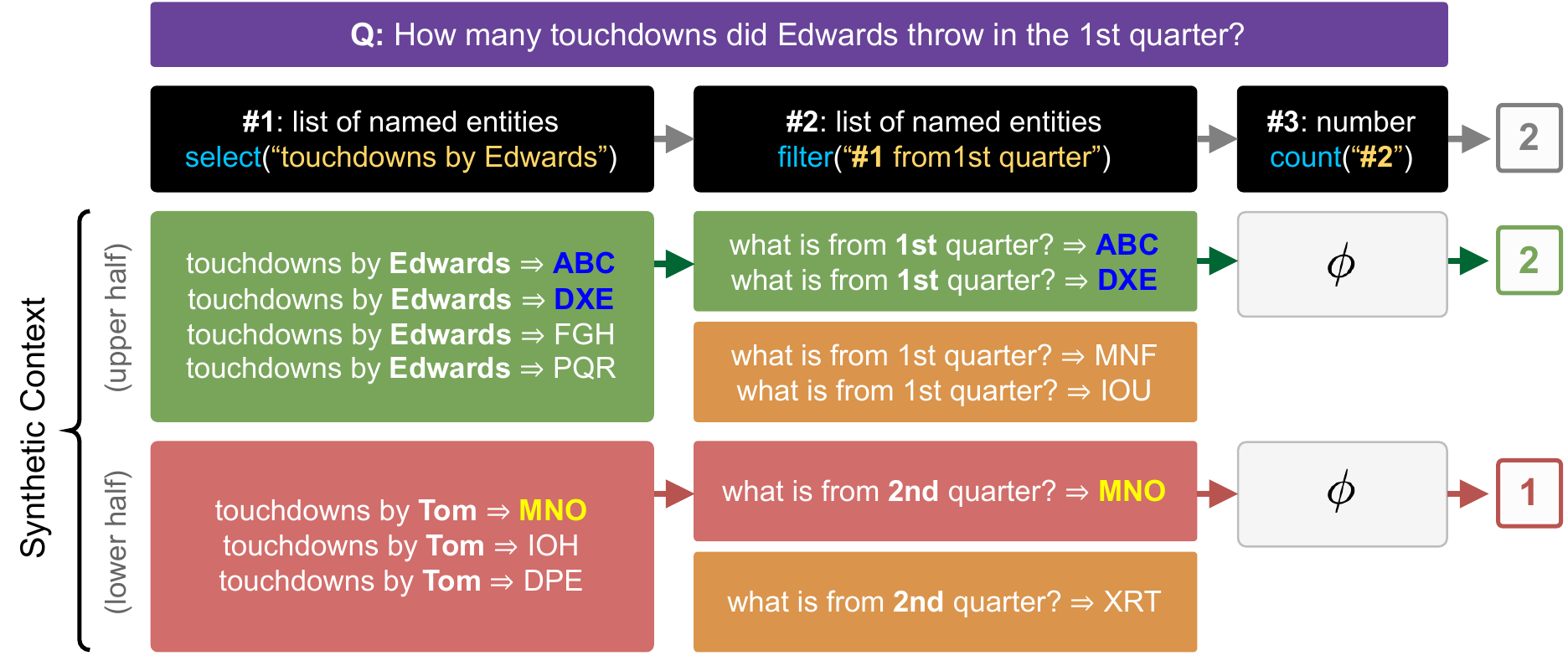}
	\caption{A simplified example of a QA instance in \teabreac, with a (simplified) real question from the DROP dataset and the synthetic context we construct for it using the question's 3-step decomposition. Statements in red, yellow and green form the synthetic context. The instance satisfies desirable properties \textbf{P1}, \textbf{P2}, and \textbf{P3}, and thus helps robustly teach multi-step reasoning skills.}
	\label{fig:synthetic-instance-example}
\end{figure*}

These primitives don't always have a clearly defined output type. While in most cases the output type is obvious (e.g., \texttt{arithmetic\_sum} returns a number), for some of them (\texttt{select}, \texttt{project}, \texttt{filter}), it's under-defined. E.g., \texttt{select("number of soldiers in USA")} should output a number, \texttt{select("when did India get independence")} should output a date, and \texttt{select("countries surrounding India")} should output a list of named entities. For such primitives, we use heuristic rules and type propagation on the global structure of $P$ to infer expected types and structures of output. We call the program having type information for each step a \textbf{Typed Program} $\tilde{P}$, an example of which is shown in Fig.~\ref{fig:qdmr-to-typed-program}.\footnote{Since these programs are more precise and executable, future work may also use them to design better explicit multi-step reasoning systems~\cite{tmn}.}

\subsubsection{Synthetic Context + Answer:}

Next, we generate $C$ and $A$ from the typed program $\tilde{P}$. We generate $C$ by grounding the \textbf{predicates} derived from the QDMR $D$ with random \textbf{entities}. Fig.~\ref{fig:synthetic-instance-example} shows an example of $C$ for a program with three steps. \textbf{Predicates}: The predicates that need to be grounded belong to four primitives, i.e., \texttt{select}, \texttt{project}, \texttt{filter}, \texttt{boolean}. Example in Fig.~\ref{fig:synthetic-instance-example} uses \texttt{select} and \texttt{filter}. Examples involving \texttt{project} and \texttt{boolean} are shown in App.~\ref{sec:more-multihop-examples}. \textbf{Entities:} The grounded entities are of 3 types: number, date, or named entity\footnote{Numbers are in 0 to 1 million, dates are from year 1100 to 2022, and named entities are any sequence of 3 letters.}. Since our programs are typed, we know which predicate should be grounded with which entity type. E.g., \texttt{select("number of soldiers in USA")} should be grounded with a number.

\paragraph{Minimizing reasoning shortcuts.} Naively creating $C$ using QDMR $D$ can introduce shortcuts that models can exploit and bypass the necessary reasoning. Note that QDMR is a sequence of steps where each step $s_i$ can use answers from zero or more previous steps; e.g., \texttt{``return number \#2 where \#3 is least goals"} in Fig~\ref{fig:qdmr-to-typed-program} (top) uses the answer from step \#2 and \#3. However, if there is only one player who scored field goals, all the steps can be ignored. To ensure models learn the intended reasoning, our goal is to create $C$ such that one can't bypass the intended reasoning (or program) steps and still arrive at the correct answer $A$. To this end, we ground the predicates with entities such that the following three properties hold:

\paragraph{P1: Answers to dependent steps can't be ignored.}

If step $s_j$ is dependent on step $s_i$, then the answer to $s_j$ can't be identified without knowing the answer to $s_i$. E.g., in Fig~\ref{fig:synthetic-instance-example}, step \#2 is asking \textit{``which of the touchdowns by Edward are \ul{from the first quarter}"}. Since there are many touchdowns \textit{``from the 1st quarter"}, and only some of them are \textit{``touchdowns by Edward"} (indicated in \textcolor{blue}{blue}), one can't narrow down the answer to step \#2 without knowing step \#1's answer. We ensure this property for different operators differently. E.g., for \texttt{filter}, we ensure the answer is always a \textit{proper} subset of all the entities grounded with that predicate (\{ABC, DXE\} $\subset$ \{ABC, DXE, MNF, IOU\} in Fig.~\ref{fig:synthetic-instance-example}).

\paragraph{P2: Steps can't be no-op.}

The input and output of steps can't be the same, as otherwise the reasoning in that step can be bypassed. E.g., in Fig~\ref{fig:synthetic-instance-example}, step \#2 is asking \textit{``which of the touchdowns by Edward are \ul{from the 1st quarter}"}. There are many \textit{``touchdowns by Edward"}, but only some of them are \textit{``from the 1st quarter"} (indicated in \textcolor{blue}{blue}). So, ignoring step \#2 (i.e., treating it as a no-op) would result in an incorrect answer being used for subsequent steps. We ensure this property for different operators differently. E.g., for \texttt{filter} operator, we ensure the answer to the step is always a \textit{proper} subset of the answer to the dependent step (\{ABC, DXE\} $\subset$ \{ABC, DXE, FGH, PQR\} in Fig.~\ref{fig:synthetic-instance-example}).

\paragraph{}
Properties \textbf{P1} and \textbf{P2} ensure step-by-step execution will lead to the gold answer, but there is only one possible complete execution that leads to \textit{an} answer. As a result, the question can be completely ignored. To fix this, we have a third property:

\paragraph{P3: Context also supports a different answer to a contrastive question.} 

Just as we generate facts for the gold chain of reasoning (upper half in Fig.~\ref{fig:synthetic-instance-example}), we also generate facts for distractor chain (lower half in Fig.~\ref{fig:synthetic-instance-example}), using potentially perturbed predicates (e.g., Edward $\Rightarrow$ Tom, 1st $\Rightarrow$ 2nd). This ensures there is always one minimally different (contrastive~\cite{contrastsets}) question that results in a different answer in the same context. E.g., \textit{``How many touchdowns did \ul{Tom} throw in the \ul{2nd} quarter"} results in the answer \textcolor{red}{1}, different from the gold answer \textcolor{teal}{2} in Fig.~\ref{fig:synthetic-instance-example}. To perturb predicates, we swap numbers, dates, and named entities (PERSON, ORG, etc.) with a similar entity of the same type. The cases where predicate doesn't have an entity, we use a similar but different and type-consistent predicate from a different question as a perturbed predicate. E.g., \textit{``yards of rushing touchdowns"} could be perturbed to \textit{``yards of passing touchdowns"}. To do this, we retrieve the top 30 type-consistent predicates with the highest word-overlap not exceeding 75\%, and sample one.

\paragraph{}

We note that past work of \citet{musique} has also considered similar properties to create hard-to-cheat multi-step QA datasets. Our P1 is similar to the first part of their MuSiQue condition (the 2nd part isn't needed here as artificial entities make it impossible to ignore the context). Our P2 is new and especially pertinent to \teabreac because of its list-based filter operations. Our P3 is also new and results in stronger question dependence than MuSiQue because of the emphasis on a minimally contrastive reasoning chain (as opposed to \emph{any} additional reasoning chain which a context in MuSiQue often also supports).

To construct QA instances with properties P1-P3, we iterate through the program steps maintaining the step-wise answers and distractors for gold reasoning chain (upper half of Fig.~\ref{fig:synthetic-instance-example}) and the distractor reasoning chain (lower half of Fig.~\ref{fig:synthetic-instance-example}) respectively. For steps containing grounding predicates (\texttt{select}, \texttt{filter}, \texttt{project}, \texttt{boolean}), we ground the predicate with random entities of appropriate type and cardinality as defined by typed program. While doing such groundings we make sure the aforementioned properties satisfy. The final step answer is the answer $A$ for the QA instance. The detailed description and pseudo-code to generate QA instances is given in App. \ref{sec:algorithm}.

\subsection{Dataset Generator}
\label{subsec:dataset-generator}

Now that we have a way to generate QA instance from a (question, QDMR) pair, we can generate a dataset by just using  questions from datasets with annotated QDMRs.  However, we find that the natural distribution of the \emph{reasoning patterns} in these datasets is extremely long-tailed. We define \textbf{reasoning pattern} as a unique sequence of primitives in the program. E.g., program in Fig.~\ref{fig:synthetic-instance-example} has 3 steps having \texttt{select}, \texttt{filter} and \texttt{count} primitives, so the reasoning pattern is \texttt{``select filter count"}.

Generating instances uniformly from such QDMRs would end up skewing the distribution of questions towards the popular patterns and result in the model overfitting to these patterns. To fix this, our dataset generator: (i) samples a reasoning pattern, (ii) samples a question-QDMR pair from that reasoning pattern, (iii) possibly perturbs question entities (named entities, dates, numbers, ordinals) with a closely similar entity of the same type,\footnote{
e.g., Edward $\Rightarrow$ Tom to create a new question: "How many touchdowns did Tom throw in the 1st quarter?". Since this perturbation is similar to the one used to create distractor chains, it makes distinguishing these distractor chains in the unperturbed questions from the gold chains in the perturbed questions much harder and better enforces property P3 in \S \ref{subsec:instance-generator}.} and (iv) invokes the instance generator. The resulting training dataset has about 900 reasoning patterns with the top 10 common patterns having only 4\% of examples (compared to 70\% had we not done such balancing).

\subsection{Additional QA Instances for Primitives}
\label{subsec:primitive-data-generator}

We also generate instances to teach 44 individual primitives, using simple templates similar to \citet{genbert}.  E.g., for primitive \texttt{filter\_a\_where\_b\_is\_compared\_to}, a question could be ``Entities that have value larger than 948768.92?" and context could be ``Entity AFE has value 871781. Entity RQX has value 989,517.24." resulting in the answer [`RQX']. App.~\ref{sec:primitives-list} gives example instances for all the primitives. Each primitive has 30K training and 1K development instances.

\subsection{Final Dataset}

Final \teabreac dataset has 525K and 15K train and development multi-step QA instances respectively, and has about 900 reasoning patterns. To create it we use publicly available QDMRs from QA and semantic parsing datasets, DROP~\cite{drop}, ComplexWebQuestions~\cite{complexwebqns}, HotpotQA~\cite{hotpotqa}, SPIDER~\cite{spider}, ComQA~\cite{comqa}, ATIS~\cite{atis}. We use both \texttt{low} and \texttt{high} level QDMRs limited to 2-6 reasoning steps.

\section{Experiments}
\label{sec:experiments}

\begin{table*}[htbp]
\small
\setlength{\tabcolsep}{3.8pt}
\makebox[\textwidth][c]{

\begin{tabular}{p{0.3cm}p{2.3cm}ccccccc}
\toprule
      &       & \multicolumn{5}{c}{In-distribution Evaluation} & \multicolumn{2}{c}{Robustness Evaluation} \\
\cmidrule(lr){3-7}\cmidrule(lr){8-9}
        & Model & DROP & TAT-QA & IIRC-G & IIRC-R & NumGLUE & DROP-CS & DROP-BPB\\

        \midrule
        & Bart-L                  &  \ph{72.3} $|$ \ph{73.3} & \ph{44.8} $|$ \ph{43.9} & \ph{66.9} $|$ \ph{65.0} & \ph{44.8} $|$ \ph{41.7} & \ph{46.0} $|$ \ph{41.9} & \ph{53.7} & \ph{51.5} \\
        \rowcolor{highlightclr} \cellcolor{white}
        & \ \ + \teabreac \icon   &  \pg{81.3} $|$ \pg{80.7} & \pg{54.2} $|$ \pg{53.7} & \pg{76.2} $|$ \pg{75.3} & \pg{48.5} $|$ \pg{45.6} & \pg{52.5} $|$ \pg{49.1} & \pg{61.8} & \pg{59.3} \\

        \cmidrule{2-9}
        & T5-L                    &  \ph{76.1} $|$ \ph{77.1} & \ph{47.2} $|$ \ph{46.3} & \ph{68.0} $|$ \ph{63.6} & \ph{45.4} $|$ \ph{38.9} & \ph{49.7} $|$ \ph{42.9} & \ph{53.4} & \ph{56.4} \\
        \rowcolor{highlightclr} \cellcolor{white}
        & \ \ + \teabreac \icon   &  \pg{81.4} $|$ \pg{81.1} & \pg{58.3} $|$ \pg{56.9} & \pg{72.9} $|$ \pg{72.8} & \pg{46.1} $|$ \pg{45.7} & \pg{53.3} $|$ \pg{49.8} & \pg{60.1} & \pg{63.2} \\

        \cmidrule{2-9}
        \multirow{-5}{*}{\rotatebox[origin=c]{90}{\parbox[c]{1.5cm}{\centering Plain LMs}}}
        & T5-3B                   &  \ph{82.0} $|$ \ph{82.1} & \ph{49.8} $|$ \ph{51.1} & \ph{70.9} $|$ \ph{68.2} & \ph{46.4} $|$ \ph{40.9} & \ph{54.9} $|$ \ph{49.7} & \ph{61.8} & \ph{63.6} \\
        \rowcolor{highlightclr} \cellcolor{white}
        & \ \ + \teabreac \icon   &  \pg{86.7} $|$ \pg{86.5} & \pg{65.5} $|$ \pg{63.8} & \pgb{78.6} $|$ \pgb{79.5} & \pgb{52.5} $|$ \pgb{51.0} & \pgb{57.3} $|$ \pgb{54.3} & \pgb{66.8} & \pgb{69.7} \\

        \midrule
        & NT5-S                   &  \ph{72.7} $|$ \ph{73.0} & \ph{51.9} $|$ \ph{51.9} & \ph{71.3} $|$ \ph{71.4} & \ph{45.2} $|$ \ph{44.3} & \ph{37.0} $|$ \ph{32.7} & \ph{46.4} & \ph{51.8} \\
        \rowcolor{highlightclr} \cellcolor{white}
        & \ \ + \teabreac \icon   &  \pg{75.1} $|$ \pg{75.3} & \pg{53.4} $|$ \pg{52.8} & \pr{70.4} $|$ \pr{70.3} & \pr{44.9} $|$ \pr{44.2} & \pg{50.7} $|$ \pg{47.5} & \pg{52.9} & \pg{54.2} \\

        \cmidrule{2-9}
        & PReasM-L                &  \ph{80.0} $|$ \ph{80.2} & \ph{48.7} $|$ \ph{49.7} & \ph{74.5} $|$ \ph{73.3} & \ph{45.5} $|$ \ph{40.9} & \ph{52.3} $|$ \ph{46.4} & \ph{57.3} & \ph{56.1} \\
        \rowcolor{highlightclr} \cellcolor{white}
        & \ \ + \teabreac \icon   &  \pg{83.2} $|$ \pg{83.4} & \pg{61.7} $|$ \pg{60.4} & \pg{77.2} $|$ \pg{77.9} & \pg{50.5} $|$ \pg{47.6} & \pg{53.1} $|$ \pg{49.2} & \pg{60.8} & \pg{64.4} \\

        \cmidrule{2-9}
        & POET-L                  &  \ph{79.6} $|$ \ph{79.4} & \ph{52.8} $|$ \ph{53.1} & \ph{71.8} $|$ \ph{73.8} & \ph{47.5} $|$ \ph{44.3} & \ph{50.7} $|$ \ph{45.5} & \ph{58.3} & \ph{55.6} \\
        \rowcolor{highlightclr} \cellcolor{white}
        \multirow{-7}{*}{\rotatebox[origin=c]{90}{\parbox[c]{2.4cm}{\centering Numerate LMs}}}
        & \ \ + \teabreac \icon   &  \pg{82.2} $|$ \pg{82.1} & \pg{55.6} $|$ \pg{54.1} & \pg{76.8} $|$ \pg{76.0} & \pg{49.1} $|$ \pg{46.6} & \pg{53.4} $|$ \pg{50.2} & \pg{64.0} & \pg{60.7} \\

        \midrule
        & Best Published & \pt{$^a$}\ph{87.5} $|$ \ph{87.8}$^a$ & \ph{$^b$}\ph{81.3} $|$ \ph{78.0}$^b$ & \pt{$^c$}\ph{77.4} $|$ \ph{75.0}$^c$ & \pt{$^d$}\ph{50.6} $|$ \ph{50.5}$^d$ & \pt{$^e$}\ph{n / a} $|$ \ph{48.8}$^e$ & \pt{$^f$}\ph{54.2}$^f$ & \pt{$^g$}\ph{65.9}$^g$ \\

        \bottomrule
\end{tabular}
}

\caption{F1 scores of in-distribution and robustness evaluation of language models (LMs) with and without \icon \xspace \teabreac pretraining on dev and test sets. Pretraining LMs on \teabreac improves their in-distribution performance and robustness across multiple QA datasets, for both plain and numerate LMs. In-distribution evaluation scores are (dev $|$ test) scores. Robustness evaluations are on test-only contrast sets. The suffixes `-3B', `-L' and `-S' refer to model sizes 3B, large and small, respectively.
\pg{Green} (underlined) indicates \teabreac pretraining improves the underlying model's performance, while \pr{red} (not underlined) indicates it does not. \bx{Bold} indicates that the \teabreac-pretrained model sets a new state of the art among published models. EM scores are provided in Appendix~\ref{sec:results-in-em}. $a:$ \citet{drop-sota}, $b:$ \citet{drop-sota}, $c:$ \citet{preasm}, $d:$ \citet{mitigating-retrieval-negs}, $e:$ \citet{numglue}, $f:$ \citet{contrastsets}, $g:$ \citet{bpb}.}
\label{table:main-results-in-f1}
\end{table*}

To test the effectiveness of \teabreac pretraining, we compare models directly fine-tuned on target datasets with models first pretrained on \teabreac\footnote{Models work well on \teabreac (see App.~\ref{sec:teabreac-results}).} before fine-tuning.

\paragraph{Datasets.}
We evaluate \textbf{in-distribution performance} using DROP~\cite{drop}, TAT-QA~\cite{tatqa}, IIRC~\cite{iirc}, and \numglue~\cite{numglue}. For IIRC, we consider two settings: IIRC-G uses only gold supporting sentences as context while IIRC-R uses paragraphs obtained using a retrieval marginalization method~\cite{mitigating-retrieval-negs}. We evaluate \textbf{robustness} using the DROP contrast set~\cite{contrastsets} and the DROP BPB contrast set~\cite{bpb}\footnote{We use the human validated set. We also remove yes/no questions from it as DROP does not contain yes/no questions but \teabreac does, and hence it unfairly favors \teabreac pretrained models.}. To do this, we directly evaluate DROP fine-tuned models on contrast sets.


\paragraph{Models.}
We evaluate \teabreac pretraining on two kinds of (language) models or LMs. For \textbf{Plain LMs}, we use T5-Large~\cite{t5} and Bart-Large~\cite{bart}. For \textbf{Numerate LMs}---those pretrained to perform numeric reasoning via different approaches---we use NT5~\cite{nt5} based on T5-Small,\footnote{NT5 is only available in small size.} \preasm~\cite{preasm} based on T5-Large, and \poet~\cite{poet} based on BART-Large. Henceforth, we will use suffixes ``-S'', ``-L'', and ``-3B'' to refer to the model size.

We use author-provided checkpoints as our initial models and then fine-tune on the target datasets. Following NT5 and \poet, we use character tokenization in all considered models during the fine-tuning stage. In some cases, prior work has also performed similar experiments (with different implementations and hyper-parameters) that we report in App.~\ref{sec:ours-vs-others-numbers} for completeness.\footnote{Models with \teabreac pretraining outperform both our and previously reported fine-tuning implementations.} Our models are implemented using PyTorch~\cite{pytorch}, Huggingface Transformers~\cite{transformers}, and AllenNLP~\cite{allennlp}. \S\ref{sec:implementation-details} includes implementation details and training hyperparameters.

\subsection{Results}

\subsubsection*{\teabreac improves model performance}

In-distribution evaluation in Table~\ref{table:main-results-in-f1} compares performance on DROP, TAT-QA, IIRC-G, IIRC-R and \numglue. For all considered plain language models (Bart-L, T5-L, and T5-3B), \teabreac pretraining results in substantial improvements across all datasets --- 4-7 F1 points on DROP, 10-13 F1 points on TAT-QA, 10-11 F1 points on IIRC-G, 4-10 F1 points on IIRC-R, and 5-7 F1 points on \numglue. For numerate language models (NT5-S, \preasm-L, and \poet-L), \teabreac pretraining improves performance by 2-3 F1 points on DROP, 1-11 F1 points on TAT-QA, and 3-15 F1 points on \numglue. \teabreac pretraining doesn't improve \ntfive-S performance on \iircg and \iircr, but it improves \preasm-L and \poet-L performances on both datasets by 2-7 F1 points. The fact that \teabreac further improves models already pretrained on previous synthetic datasets highlights its complementarity.

Furthermore, \teabreac-pretrained T5-3B achieves new state-of-the-art performance relative to the best previously published results on \iircg, \iircr, and \numglue, reported in the last row of Table~\ref{table:main-results-in-f1}. Moreover, even the smaller \teabreac-pretrained \preasm-L and \poet-L models improve over previously published numbers on \iircg and \numglue respectively. On DROP and TAT-QA, specialized architectures (with special task-specific modules) developed for those datasets outperform \teabreac-pretrained models.

Since numerate LMs are derived by pretraining plain LMs on respective synthetic datasets, we can also directly compare such pretraining approaches with \teabreac pretraining. From Table~\ref{table:main-results-in-f1}, we can also see that the T5-L\,+\,\teabreac model is better than the \preasm-L model (T5-L pretrained on \preasm data), and the Bart-L\,+\,\teabreac model is better than the \poet-L model (Bart-L pretrained on \poet data). See App.~\ref{sec:ours-vs-others-direct-comparison} for additional comparisons.

\subsubsection*{\teabreac improves model robustness}

We evaluate the robustness in Table~\ref{table:main-results-in-f1} by comparing performance on the DROP contrast set and the DROP BPB set. For all plain language models, T5-L, T5-3B and Bart-L, \teabreac pretraining shows substantial improvements in robustness --- 5-8 F1 points improvements on DROP contrast set and on DROP BPB set. For numerate LMs, \ntfive-S, \preasm-L and \poet-L, \teabreac pretraining results in 4-7 F1 points of improvement on DROP contrast set and 2-8 points of improvement on DROP BPB set.

\subsubsection*{\icon\xspace improves more on more complex questions}

We further investigate how the improvements provided by \teabreac vary based on the complexity or number of steps of the question. To obtain the number of reasoning steps, we use our programs.\footnote{Our programs may differ in the number of steps than the source QDMR due to additional normalization and processing.} But since QDMRs, and as a result programs, are not available for all the questions, we use the number of reasoning steps in predicted programs (using a T5-Large model trained on the BREAK dataset followed by conversion into our typed programs).

\begin{figure*}[t]
    \centering
	\includegraphics[width=0.95\textwidth]{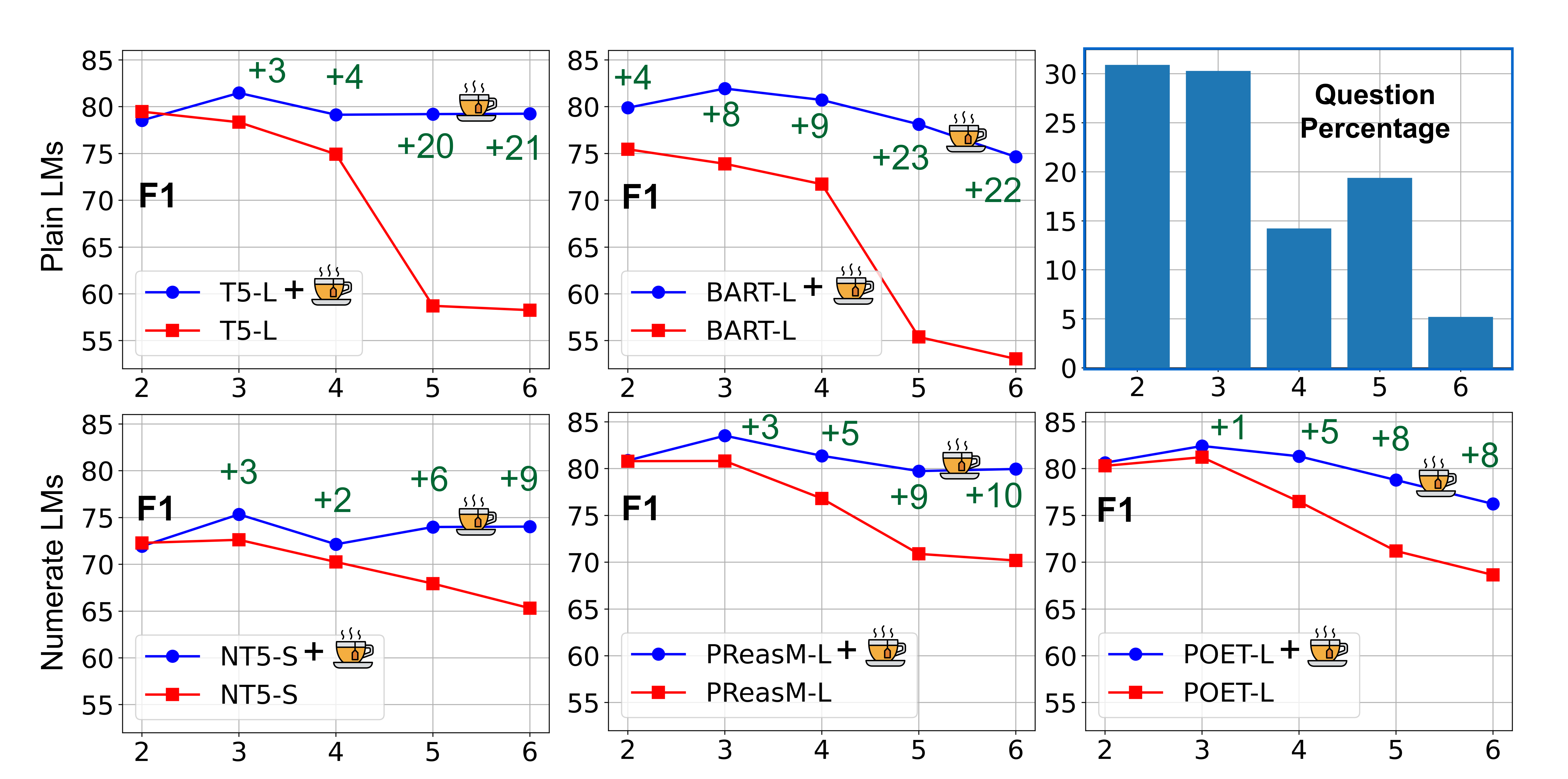}
	\caption{F1 scores for plain and numerate LMs with and without \icon \xspace \teabreac pretraining on DROP across varying numbers of steps, as determined by our programs. \teabreac pretraining helps more on more complex questions. The effect is more prominent on plain LMs like T5-L than on numerate LMs like \preasm-L. (Top Right) Histogram of percentage of questions for each step count. Because more complex questions are less frequent, improvements by \teabreac pretraining don't show up as well on the average metric for the entire dataset.}
	\label{fig:improvements-plot-by-complexity-full}
\end{figure*}

Figure~\ref{fig:improvements-plot-by-complexity-full} compares the performance of \teabreac pretraining on questions with increasing (estimated) number of steps. While the T5-L baseline model drops significantly from 79 to 58, T5-L with \teabreac pretraining stays mostly invariant to the number of steps. We thus observe a significantly larger improvement for more complex questions, where the original T5-L model struggles (e.g., 20 points gain on 4+ steps vs.\ 5 points gain on average). Similarly, for the numeracy-aware language model \preasm-Large, we see more improvement on more complex questions (e.g., 9-10 points on 4+ steps, 3.2 points on average). We see similar trends for the other models as well.

We also observe that more complex questions are much less frequent in the DROP development set (e.g., 4+ steps constitute only 25\%). This makes our large gains on more complex questions not quite visible in the aggregate metric (Table~\ref{table:main-results-in-f1}).

\subsubsection*{\teabreac Ablations}
\label{subsec:teabreac-ablations}

\begin{figure}[t]
    \centering
	\includegraphics[width=0.48\textwidth]{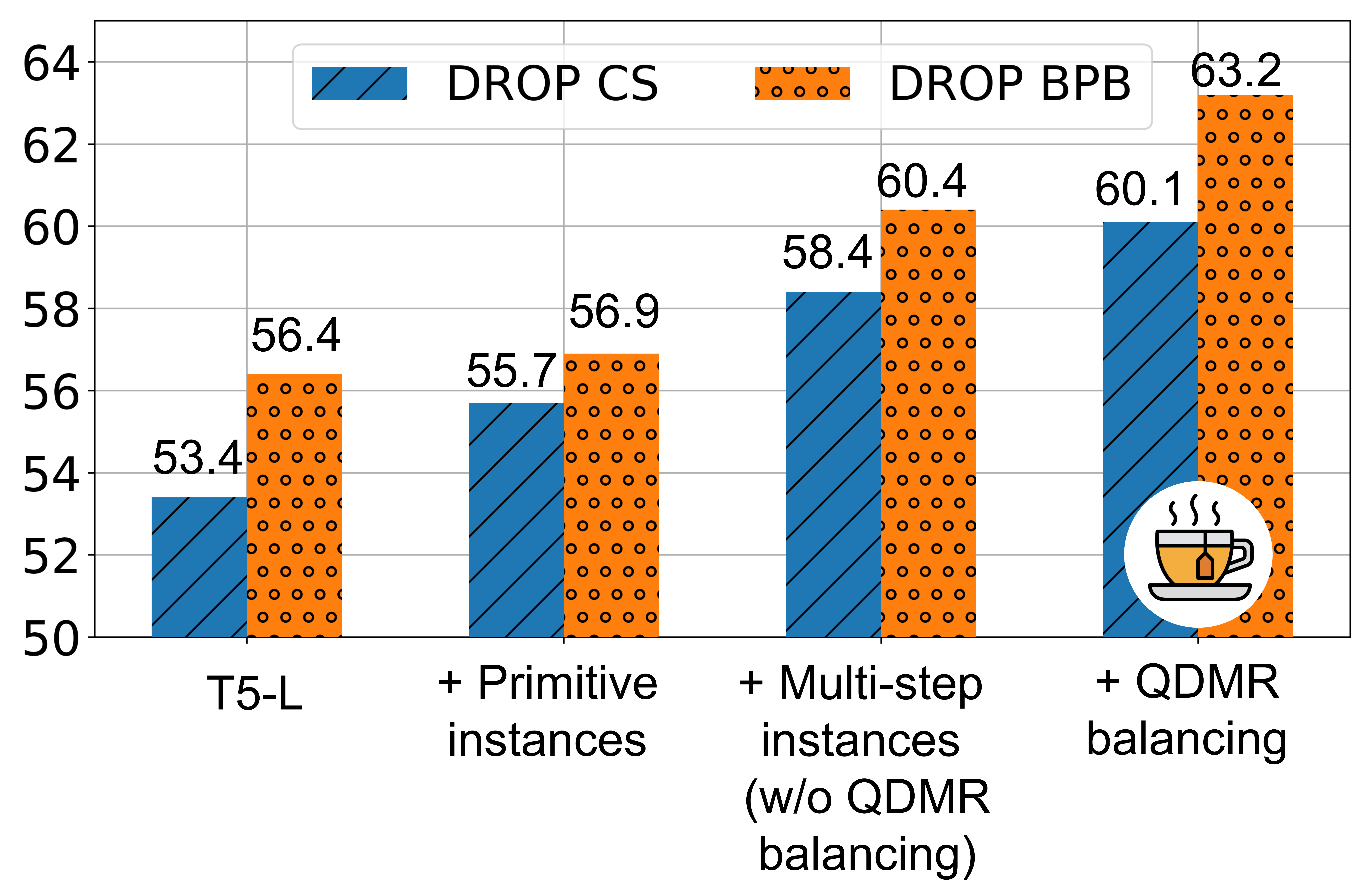}
	\caption{\icon \xspace \teabreac Ablations: All aspects of \teabreac pretraining data contribute to the overall performance: (i) primitive QA instances (ii) multi-step QA instances (iii) balancing of QDMR distribution.}
	\label{fig:teabreac-ablations-drop-cs-bpb}
\end{figure}

To assess the contribution of various aspects of \teabreac to the overall performance, we perform ablation experiments with T5-L on the DROP dataset. Fig.~\ref{fig:teabreac-ablations-drop-cs-bpb} shows the results for DROP contrast set and BPB set. Pretraining on just primitive QA instances helps by 0.5-2.3 points, which further improves by 2.7-3.5 points when adding multi-step QA instances without QDMR-balancing (\S~\ref{subsec:dataset-generator}). Finally, if we add multi-step instances with QDMR-balancing instead, we get an additional 1.7-2.8 points of improvement. DROP development set has similar trends but with lower absolute differences, potentially due to shortcuts (see App.~\ref{sec:teabreac-ablations-drop-dev}).

\section{Conclusions}

Despite large LMs' impressive reading abilities and the availability of large scale multi-step QA datasets requiring a rich set of reasoning skills, LM-based QA models do not reliably learn to use such skills for answering complex questions. In this work, we show that the greater control that synthetic contexts offer can be leveraged to create a teaching dataset where models \emph{can} learn a broad range of reasoning skills in a reliable manner, especially for more complex questions.

Our transfer results from synthetic data to actual QA datasets add to the growing line of work that shows synthetic datasets can in fact be used to inject useful skills are valuable for real, natural language tasks. Given the artifact issues in real datasets (specifically, in their contexts) and the difficulty in controlling for them via perturbations, we present a viable alternative: leveraging existing multi-step questions for their broad reasoning patterns but using synthetic contexts for carefully constructing teaching datasets, where models can learn the \textit{right way} to reason.

\section*{Ethical Considerations}

The source dataset that \teabreac is created from, i.e., BREAK, is publicly available with the MIT license which allows us to modify and release the dataset. \teabreac models and datasets are released under the CC BY 4.0 License.\footnote{\url{https://creativecommons.org/licenses/by/4.0}}

Since \teabreac uses questions and decompositions from existing datasets, it may also inherit the social biases present in these underlying datasets. We haven't taken any explicit steps to remove such potential biases as it's not in the scope of this work. But we advise the users of the \teabreac dataset and models to take appropriate caution if deploying them in any real user-facing application.
\section*{Limitations}

We proposed a pretraining approach to teach a broad range of multi-step reasoning skills to language models. Even though such pretraining doesn't have to be repeated for each target dataset, there is a significant computational cost to pretraining. E.g., our T5-Large pretraining takes 5 days on a RTX A6000 GPU. This is precisely the reason why we haven't conducted experiments with even larger models such as T5-11B. Identifying more compute-efficient ways to teach models such skills remains an interesting open problem.

In general, we expect \teabreac pretraining to improve downstream performance on the datasets that require complex aggregation operations and diverse compositions of them. We have shown the effectiveness of \teabreac pretraining on several multi-step QA datasets which fit this criteria. However, this is not the case for other multi-step QA datasets like QASC, HotpotQA, 2WikiMultihopQA, and MuSiQue, which involve simpler compositions. \teabreac pretraining thus may not lead to similar gains on these datasets. More broadly, multi-step QA datasets we have considered form only a small subset of the vast number of QA and NLU tasks the NLP community is interested in. It's possible that \teabreac pretraining is unhelpful and even harmful to the performance of LMs on these other tasks where our learned multi-step skills are not as relevant, such as commonsense understanding.



The skills taught in \teabreac are limited by the skills captured (or capturable) by QDMRs. While expanding the scope of QDMR operators and the datasets annotated with them can automatically expand the scope of \teabreac, the current approach is still limited to datasets where one can easily define and obtain QDMRs.

Lastly, while \teabreac enables the teaching of reasoning skills to any text-to-text model, these black-box models don't provide explanations, making it hard to analyze their underlying reasoning. Hence, we are unable to check whether models trained on it are necessarily performing the required multi-step reasoning. We only provide indirect empirical evidence by evaluations on contrast sets. 



\subsection*{Acknowledgments}

The authors thank the reviewers for their valuable feedback and suggestions, and Ori Yoran and Qian Liu for providing the \preasm and \poet models, respectively. This work was supported in part by the National Science Foundation under grants IIS-2007290 and CCF-1918225.

\bibliography{references}
\bibliographystyle{acl_natbib}

\clearpage
\appendix

\section{Related Work}
\label{app:related-work}

\paragraph{Question Decomposition.}

Several recent multi-step QA datasets come with question decompostion annotations~\cite{qasc,complexwebqns,strategyqa,musique,commaqa}. These works have enabled the development of \textit{explicit} multi-step reasoning systems that first decomposes a question into sub-questions, and answers the sub-questions step-by-step to arrive at the answer~\cite{decomprc,tmn,musique,patel2022question,khot2022decomposed}. In contrast, our goal is to use decompositions to teach language models multi-step reasoning implicitly (within the model).

Since each dataset has its own decomposition format, they have led to narrow dataset-specific solutions. In contrast, the BREAK dataset~\cite{break} defined a standardized format for several QA datasets. So in this work, we use them to build a teaching dataset for broad reasoning skills.

\paragraph{Robust Multi-step Reasoning.}
Past work has shown how to perturb existing multi-step QA instances to prevent shortcuts and incentivize robust reasoning. \citet{jiang2019avoiding,ding2021reasoning} created adversarial multi-step question by perturbing the reasoning chains in HotpotQA~\cite{hotpotqa}. Other datasets~\cite{dire,musique,multihop-robustifying} incentivize robustness via minimally perturbed unanswerable questions. Our approach targets a broader set of questions and eliminates multiple reasoning shortcuts. 

The closest work to ours is the Break-Perturb-Build (BPB) dataset~\cite{bpb}. BPB also uses QDMR but to create contrastive \textit{questions} via small question perturbation~\cite{learningdifference,contrastsets}. Unlike us, they use the existing context with reasoning shortcuts that can be hard to eliminate with only question perturbation (e.g., no distractors). Additionally it is mainly used for evaluation (as we also do) and hasn't been shown to improve models by training on it.

\paragraph{Data Augmentation for QA.}

Several past works have used data augmentation via synthetic datasets to improve QA performance. Following works are most relevant to our approach.  ~\citet{genbert} created a synthetic dataset using a few hand-crafted templates for injecting numerical reasoning skills (along with a specialized architecture). This dataset was also later used to build a numeracy-aware T5~\cite{t5} model: NT5~\citet{nt5}. ~\citet{preasm} created a synthetic dataset using 13 handcrafted multi-step QA reasoning patterns applied on wikipedia tables. Lastly, ~\citet{poet} showed that pretraining language models on synthetic dataset derived from input and output of program executors (arithmetic, logic-based and SQL-based) can also improve downstream QA performance. In contrast to these works, we use actual questions from a wide range of real datasets to teach a broad range of multi-step reasoning skills. 

\section{Algorithm to Generate QA Instances}
\label{sec:algorithm}

Algorithm \ref{alg:instance_gen} shows the pseudo-code for generating QA instances satisfying the three properties discussed in \S~\ref{sec:construction}. The \texttt{GenQAInstance} function takes question Q, QDMR D and expected answer cardinality \texttt{N} of the answer, and attempts to generate a QA instance with desirable properties for 200 maximum tries. For a given question, QDMR pair, we vary \texttt{N} $\in \{1,2,3,4\}$. The \texttt{facts} represent list of grounded predicates that form the context, \texttt{state.ans} represents stepwise answers for gold reasoning chain (e.g., green boxes in Fig.~\ref{fig:synthetic-instance-example}), and \texttt{state.dis} represents stepwise answers for distractor reasoning chain (e.g., red boxes in Fig.~\ref{fig:synthetic-instance-example}). These are initialized to $\emptyset$(\texttt{L3}) and updated during the instance generation.

To construct a QA instance, we iterate through the program (or QDMR) steps. For each step, we create facts for the gold reasoning chain by grounding the predicate in the QDMR and update the facts and answer state accordingly using the \texttt{execute} function. E.g., in step \#2 in Fig.~\ref{fig:synthetic-instance-example}, the facts in the top-half are added and $\{$ABC, DXE$\}$ is marked as the current answer state. The \texttt{execute} function will generate these facts and answers such that properties P1 and P2 are satisfied or return False if it can't. We similarly generate facts and update the state for the distractor reasoning chain (\texttt{L7)} by using a perturbed (\texttt{L6}) QDMR predicate (e.g., Edward $\Rightarrow$ Tom, 1st $\Rightarrow$ 2nd in Fig.~\ref{fig:synthetic-instance-example}). This generates the facts and reasoning chain shown in the lower half of  Fig.~\ref{fig:synthetic-instance-example} ensuring property P3 is satisfied.

The implementation of \texttt{execute} function is dependent on the program primitives (Table~\ref{tab:primitives}) and will be provided in the released code. But broadly speaking there are two classes of primitives: (1) primitives like \texttt{select} and \texttt{filter} that need to first add facts by grounding the predicate, and then update the answer state for that step (e.g., step \#1 and \#2 in Fig.~\ref{fig:synthetic-instance-example}) (2) primitives like \texttt{count} with no additional grounding of facts and only need to update the state based on the underlying computation (e.g., step \#3 in Fig.~\ref{fig:synthetic-instance-example}).

\begin{algorithm}[t!]
\caption{Pseudo-code for generating QA instances from question Q, QDMR D, and answer cardinality N}
    \label{alg:instance_gen}
\begin{footnotesize}
\begin{algorithmic}[1]
\renewcommand{\algorithmicindent}{1em}
\renewcommand{\algorithmiccomment}[1]{\hfill$ \triangleright$ \textit{#1}}
\Function{GenQAInstance}{
Q, D, N}
\For{$1 \leq i \leq 200$} \Comment{Max retries}
\State state.ans, state.dis, facts $\gets$ $\emptyset$
\For{step $\in$ qdmr.steps}
\State ans\_succ $\gets$ execute(step, state.ans, facts) \Comment{Update for gold reasoning chain}
\State maybe\_perturb(step) \Comment{Perturb predicate for distractor chain}
\State dis\_succ $\gets$ execute(step, state.dis, facts) 
\Comment{Update for distractor reasoning chain}
\If{\textbf{not} ans\_succ \textbf{or} \textbf{not} dis\_succ}
 \State failed $\gets$ True
 \State break
\EndIf
\EndFor
\If{(\textbf{not} failed \textbf{and} \\
\hspace{3em} accept(state, facts, ans\_num))}
  \State return QA(Q, \Comment{question} \\
  \hspace{7em} facts,  \Comment{context}\\
  \hspace{7em} state.ans[-1]) \Comment{gold answer}
\EndIf
\EndFor
\EndFunction
\end{algorithmic}
\end{footnotesize}
\end{algorithm}

If all the steps finish with success, we check if the generation is \texttt{acceptable} (\texttt{L14}) before creating a QA instance. For it to be acceptable, the generated answer cardinality must match the expected value, the number of facts must be within 25, and the final answer for gold and distractor reasoning chains must be different. We create a reading comprehension QA instance with the input question Q as question, facts as the context (concatenated after shuffling), and the answer at the final step as the gold answer. 

\section{Our Implementation vs Previously Reported Numbers}
\label{sec:ours-vs-others-numbers}

To test the effectiveness of \teabreac pretraining, we compare models directly fine-tuned on target datasets with models first pretrained on \teabreac and then fine-tuned on target datasets. For a fair comparison of the fine-tuning experiments, we do the direct fine-tuning on the target datasets using our implementation instead of relying of previously reported numbers which may have other differences. Moreover, previously reported numbers are only sparsely available across the model-dataset pairs we consider, which is another reason to use our implementation. Table~\ref{table:our-vs-others-results} shows results obtained by our implementation vs results reported by prior works (NT5~\cite{nt5}, \preasm~\cite{preasm} and POET~\cite{poet}), where available. Irrespective of implementation, models with \teabreac outperform prior approaches.

Note that following~\citet{nt5} and~\citet{poet}, we employ character tokenization for numbers, but it wasn't employed by~\citet{preasm}. Therefore, our results obtained by our implementation are significantly better than the ones reported in~\citet{preasm} for DROP, where numerical reasoning is crucial.

\begin{table*}[htbp]
\small
\setlength{\tabcolsep}{3.6pt}
\makebox[\textwidth][c]{

\begin{tabular}{p{0.3cm}p{3.95cm}ccccccc}
\toprule
      &       & \multicolumn{5}{c}{In-distribution Evaluation} & \multicolumn{2}{c}{Robustness Evaluation} \\
\cmidrule(lr){3-7}\cmidrule(lr){8-9}
        & Model & DROP & TAT-QA & IIRC-G & IIRC-R & NumGLUE & DROP-CS & DROP-BPB\\

        \midrule
        & Bart-L {\footnotesize (\citet{poet})}       &  \ph{69.2} $|$ --\pt{x.x}& \ph{46.7} $|$ --\pt{x.x}&           ---           &           ---           &           ---           &    ---    &   ---     \\
        & Bart-L {\footnotesize (our implemented)} &  \ph{72.3} $|$ \ph{73.3} & \ph{44.8} $|$ \ph{43.9} & \ph{66.9} $|$ \ph{65.0} & \ph{44.8} $|$ \ph{41.7} & \ph{46.0} $|$ \ph{41.9} & \ph{53.7} & \ph{51.5} \\
        \rowcolor{highlightclr} \cellcolor{white}
        & \ \ + \teabreac \icon     &  \pg{81.3} $|$ \pg{80.7} & \pg{54.2} $|$ \pg{53.7} & \pg{76.2} $|$ \pg{75.3} & \pg{48.5} $|$ \pg{45.6} & \pg{52.5} $|$ \pg{49.1} & \pg{61.8} & \pg{59.3} \\

        \cmidrule{2-9}
        \multirow{-3.5}{*}{\rotatebox[origin=c]{90}{\parbox[c]{1.5cm}{\centering Plain LMs}}}
        & T5-L {\footnotesize (\citet{preasm})}       &  \ph{64.6} $|$ \ph{65.0} &           ---           & \ph{69.9} $|$ \ph{67.1} & \ph{47.4} $|$ \ph{41.0} &           ---           &    ---    &   ---     \\
        & T5-L {\footnotesize (our implemented)}   &  \ph{76.1} $|$ \ph{77.1} & \ph{47.2} $|$ \ph{46.3} & \ph{68.0} $|$ \ph{63.6} & \ph{45.4} $|$ \ph{38.9} & \ph{49.7} $|$ \ph{42.9} & \ph{53.4} & \ph{56.4} \\
        \rowcolor{highlightclr} \cellcolor{white}
        & \ \ + \teabreac \icon     &  \pg{81.4} $|$ \pg{81.1} & \pg{58.3} $|$ \pg{56.9} & \pg{72.9} $|$ \pg{72.8} & \pr{46.1} $|$ \pg{45.7} & \pg{53.3} $|$ \pg{49.8} & \pg{60.1} & \pg{63.2} \\

        \midrule
        & NT5-S {\footnotesize (\citet{nt5})}         &  \ph{70.3} $|$ \ph{70.8} &           ---           &           ---           &           ---           &           ---           &    ---    &   ---     \\
        & NT5-S {\footnotesize (our implemented)} &  \ph{72.7} $|$ \ph{73.0} & \ph{51.9} $|$ \ph{51.9} & \ph{71.3} $|$ \ph{71.4} & \ph{45.2} $|$ \ph{44.3} & \ph{37.0} $|$ \ph{32.7} & \ph{46.4} & \ph{51.8} \\
        \rowcolor{highlightclr} \cellcolor{white}
        & \ \ + \teabreac \icon     &  \pg{75.1} $|$ \pg{75.3} & \pg{53.4} $|$ \pg{52.8} & \pr{70.4} $|$ \pr{70.3} & \pr{44.9} $|$ \pr{44.2} & \pg{50.7} $|$ \pg{47.5} & \pg{52.9} & \pg{54.2} \\

        \cmidrule{2-9}
        & PReasM-L {\footnotesize (\citet{preasm})}      &  \ph{72.3} $|$ \ph{72.6} &           ---           & \ph{77.4} $|$ \ph{75.0} & \ph{50.0} $|$ \ph{45.1} &           ---           &    ---    &   ---     \\
        & PReasM-L {\footnotesize (our implemented)} &  \ph{80.0} $|$ \ph{80.2} & \ph{48.7} $|$ \ph{49.7} & \ph{74.5} $|$ \ph{73.3} & \ph{45.5} $|$ \ph{40.9} & \ph{52.3} $|$ \ph{46.4} & \ph{57.3} & \ph{56.1} \\
        \rowcolor{highlightclr} \cellcolor{white}
        & \ \ + \teabreac \icon        &  \pg{83.2} $|$ \pg{83.4} & \pg{61.7} $|$ \pg{60.4} & \pg{77.2} $|$ \pg{77.9} & \pg{50.5} $|$ \pg{47.6} & \pg{53.1} $|$ \pg{49.2} & \pg{60.8} & \pg{64.4} \\

        \cmidrule{2-9}
        & POET-L {\footnotesize (\citet{poet})}        &  \ph{80.6} $|$ --\pt{x.x}& \ph{49.6} $|$ --\pt{x.x}&           ---           &           ---           &           ---           &    ---    &   ---     \\
        & POET-L  {\footnotesize (our implemented)} &  \ph{79.6} $|$ \ph{79.4} & \ph{52.8} $|$ \ph{53.1} & \ph{71.8} $|$ \ph{73.8} & \ph{47.5} $|$ \ph{44.3} & \ph{50.7} $|$ \ph{45.5} & \ph{58.3} & \ph{55.6} \\
        \rowcolor{highlightclr} \cellcolor{white}
        \multirow{-10}{*}{\rotatebox[origin=c]{90}{\parbox[c]{3.4cm}{\centering Numerate LMs}}}
        & \ \ + \teabreac \icon      &  \pg{82.2} $|$ \pg{82.1} & \pg{55.6} $|$ \pg{54.1} & \pg{76.8} $|$ \pg{76.0} & \pg{49.1} $|$ \pg{46.6} & \pg{53.4} $|$ \pg{50.2} & \pg{64.0} & \pg{60.7} \\

        \bottomrule
\end{tabular}
}
\caption{Comparison of: (i) Results reported by prior works (NT5~\cite{nt5}, \preasm~\cite{preasm} and \poet~\cite{poet}) where available (ii) Results obtained from our implementation (iii) Results obtained by our implementation with \icon \xspace \teabreac pretraining. Irrespective of implementation, models with \teabreac outperform prior approaches. In-distribution evaluation scores are (dev $|$ test) scores. Robustness evaluations are on test-only contrast sets. The scores are in terms of F1 metric. \pg{Green} (underlined) indicates \teabreac pretraining improves the underlying model's performance, while \pr{red} (not underlined) indicates it does not.}
\label{table:our-vs-others-results}
\end{table*}

\section{Direct Comparison of \teabreac vs Previous Pretraining Methods}
\label{sec:ours-vs-others-direct-comparison}

Table~\ref{table:main-results-in-f1} shows that \teabreac pretraining improves both plain and numerate LMs. However, since the numerate LMs (\preasm and \poet) are derived by pretraining plain LMs on the respective synthetic datasets, we can also directly compare \teabreac pretraining with \preasm and \poet pretraining. Table~\ref{table:main-results-in-f1-alternative-format} shows this comparison. The results are limited to T5-Large and Bart-Large as \preasm and \poet checkpoints are not available in other sizes considered. We find that LM with \teabreac pretraining outperforms LM with \preasm or \poet pretraining.

\begin{table*}[htbp]
\small
\setlength{\tabcolsep}{5.5pt}
\makebox[\textwidth][c]{

\begin{tabular}{p{3.0cm}ccccccc}
\toprule
        & \multicolumn{5}{c}{In-distribution Evaluation} & \multicolumn{2}{c}{Robustness Evaluation} \\

        \cmidrule(lr){2-6}\cmidrule(lr){7-8}
        Model & DROP & TAT-QA & IIRC-G & IIRC-R & NumGLUE & DROP-CS & DROP-BPB\\

        \midrule
        T5-L                      &  \ph{76.1} $|$ \ph{77.1} & \ph{47.2} $|$ \ph{46.3} & \ph{68.0} $|$ \ph{63.6} & \ph{45.4} $|$ \ph{38.9} & \ph{49.7} $|$ \ph{42.9} & \ph{53.4} & \ph{56.4} \\

        PReasM (T5-L based)      &  \ph{80.0} $|$ \ph{80.2} & \ph{48.7} $|$ \ph{49.7} & \ph{74.5} $|$ \ph{73.3} & \ph{45.5} $|$ \ph{40.9} & \ph{52.3} $|$ \ph{46.4} & \ph{57.3} & \ph{56.1} \\

        T5-L +  \teabreac         &  \pg{81.4} $|$ \pg{81.1} & \pg{58.3} $|$ \pg{56.9} & \pr{72.9} $|$ \pr{72.8} & \pg{46.1} $|$ \pg{45.7} & \pg{53.3} $|$ \pg{49.8} & \pg{60.1} & \pg{63.2} \\


        \midrule

        Bart-L                  &  \ph{72.3} $|$ \ph{73.3} & \ph{44.8} $|$ \ph{43.9} & \ph{66.9} $|$ \ph{65.0} & \ph{44.8} $|$ \ph{41.7} & \ph{46.0} $|$ \ph{41.9} & \ph{53.7} & \ph{51.5} \\

        POET-L (Bart-L based)      &  \ph{79.6} $|$ \ph{79.4} & \ph{52.8} $|$ \ph{53.1} & \ph{71.8} $|$ \ph{73.8} & \ph{47.5} $|$ \ph{44.3} & \ph{50.7} $|$ \ph{45.5} & \ph{58.3} & \ph{55.6} \\

        Bart-L + \teabreac   &  \pg{81.3} $|$ \pg{80.7} & \pg{54.2} $|$ \pg{53.7} & \pg{76.2} $|$ \pg{75.3} & \pg{48.5} $|$ \pg{45.6} & \pg{52.5} $|$ \pg{49.1} & \pg{61.8} & \pg{59.3} \\


        \bottomrule
\end{tabular}
}
\caption{F1 scores of in-distribution and robustness evaluation of large-sized models with previous and our pretraining. Pretraining on \teabreac leads to better results than pretraining on \preasm or \poet.  \pg{Green} (underlined) indicates \teabreac pretraining leads to better results, while \pr{red} (not underlined) indicates \preasm or \poet leads to better results.
}
\label{table:main-results-in-f1-alternative-format}
\end{table*}

\section{Performance of LMs on \teabreac}
\label{sec:teabreac-results}

Since our goal is to teach models the reasoning skills in \teabreac, we assess how well models do on the \teabreac dataset. As shown in Table~\ref{table:teabreac-results}, models are able to learn both primitive and multi-step QA skills required in \teabreac. On primitives instances models get 92-98 F1, and on multi-step instances, models get 84-88 F1. We show in our experiments that these scores are good enough to make progress on real datasets. At the same time, these aren't perfect scores, demonstrating limitations of vanilla LM-based neural models. Thus, \teabreac can also serve as a benchmark to help design better multi-step models.

\begin{table}[t]
    \centering
    \normalsize
    \setlength{\tabcolsep}{2.5pt}
    \begin{tabular}{p{2.5cm}p{2.2cm}p{2.2cm}}\toprule
        Model &   \icon \xspace Primitive & \icon  \xspace Multi-step   \\
        \midrule

        Bart-L \icon      &  \quad \quad 91.8 & \quad 86.5 \\
        T5-L \icon        &  \quad \quad 93.8 & \quad 87.6 \\
        T5-3B \icon       &  \quad \quad 94.2 & \quad 88.9 \\
        \midrule
        NT5-S \icon       &  \quad \quad 98.1 & \quad 84.1 \\
        PReasM-L \icon    &  \quad \quad 94.2 & \quad 88.3 \\
        POET-L \icon      &  \quad \quad 91.5 & \quad 87.4 \\
        \bottomrule
    \end{tabular}
    \caption{F1 scores of models pretrained on \teabreac on its Primitive and Multi-step dev sets. Models learn the skills required in \teabreac during pretraining well, but achieving perfect score is challenging for vanilla LM-based neural models.}
    \label{table:teabreac-results}
\end{table}

\section{Results in Exact Match (EM) metric}
\label{sec:results-in-em}

In addition to the F1 results reported in Table~\ref{table:main-results-in-f1}, we also report the corresponding EM numbers in Table~\ref{table:main-results-in-em}. We see the same trends discussed in \S~\ref{sec:experiments}.

\begin{table*}[htbp]
\small
\setlength{\tabcolsep}{6pt}
\makebox[\textwidth][c]{

\begin{tabular}{p{0.3cm}p{2.3cm}ccccccc}
\toprule
      &       & \multicolumn{5}{c}{In-distribution Evaluation} & \multicolumn{2}{c}{Robustness Evaluation} \\
\cmidrule(lr){3-7}\cmidrule(lr){8-9}
        & Model & DROP & TAT-QA & IIRC-G & IIRC-R & NumGLUE & DROP-CS & DROP-BPB\\

        \midrule
        & Bart-L         &  \ph{69.2} $|$ \ph{70.0} & \ph{37.4} $|$ \ph{35.8} & \ph{62.4} $|$ \ph{60.5} & \ph{41.7} $|$ \ph{39.1} & \ph{44.5} $|$ \ph{40.4} & \ph{47.0} & \ph{46.7} \\
        \rowcolor{highlightclr} \cellcolor{white} 
        & \ \ + \teabreac \icon &  \pg{78.0} $|$ \pg{77.1} & \pg{45.9} $|$ \pg{43.9} & \pg{72.4} $|$ \pg{70.9} & \pg{45.2} $|$ \pg{42.5} & \pg{50.9} $|$ \pg{47.6} & \pg{52.8} & \pg{53.9} \\

        \cmidrule{2-9}
        & T5-L        &  \ph{73.2} $|$ \ph{73.9} & \ph{39.4} $|$ \ph{37.4} & \ph{64.0} $|$ \ph{59.3} & \ph{42.4} $|$ \ph{36.0} & \ph{48.2} $|$ \ph{41.3} & \ph{46.7} & \ph{51.2} \\
        \rowcolor{highlightclr} \cellcolor{white} 
        & \ \ + \teabreac \icon &  \pg{78.2} $|$ \pg{77.8} & \pg{50.4} $|$ \pg{47.8} & \pg{69.1} $|$ \pg{69.0} & \pg{43.1} $|$ \pg{42.7} & \pg{51.7} $|$ \pg{48.3} & \pg{52.7} & \pg{57.3} \\

        \cmidrule{2-9}
        \multirow{-5}{*}{\rotatebox[origin=c]{90}{\parbox[c]{1.5cm}{\centering Plain LMs}}}
        & T5-3B                   &  \ph{79.1} $|$ \ph{79.5} & \ph{42.5} $|$ \ph{43.1} & \ph{66.3} $|$ \ph{64.2} & \ph{44.0} $|$ \ph{38.0} & \ph{53.4} $|$ \ph{48.2} & \ph{54.7} & \ph{58.3} \\
        \rowcolor{highlightclr} \cellcolor{white}
        & \ \ + \teabreac \icon   &  \pg{83.9} $|$ \pg{83.8} & \pg{58.2} $|$ \pg{55.8} & \pg{74.5} $|$ \pg{75.4} & \pg{49.0} $|$ \pg{48.4} & \pg{55.8} $|$ \pg{52.8} & \pg{58.6} & \pg{64.4} \\

        \midrule
        & NT5-S         &  \ph{69.2} $|$ \ph{69.4} & \ph{44.2} $|$ \ph{42.3} & \ph{66.6} $|$ \ph{66.9} & \ph{41.9} $|$ \ph{41.7} & \ph{34.2} $|$ \ph{29.2} & \ph{38.8} & \ph{46.3} \\
        \rowcolor{highlightclr} \cellcolor{white} 
        & \ \ + \teabreac \icon &  \pg{71.7} $|$ \pg{71.7} & \pg{44.8} $|$ \pg{43.3} & \pr{65.6} $|$ \pr{65.3} & \pg{42.0} $|$ \pr{41.6} & \pg{49.3} $|$ \pg{46.0} & \pg{45.5} & \pg{48.2} \\

        \cmidrule{2-9}
        & PReasM-L        &  \ph{76.9} $|$ \ph{77.0} & \ph{40.8} $|$ \ph{41.2} & \ph{70.0} $|$ \ph{69.1} & \ph{42.1} $|$ \ph{38.1} & \ph{50.9} $|$ \ph{44.8} & \ph{49.9} & \ph{50.6} \\
        \rowcolor{highlightclr} \cellcolor{white} 
        & \ \ + \teabreac \icon &  \pg{80.1} $|$ \pg{80.1} & \pg{54.5} $|$ \pg{51.6} & \pg{73.0} $|$ \pg{72.9} & \pg{47.3} $|$ \pg{44.6} & \pg{51.5} $|$ \pg{47.7} & \pg{53.6} & \pg{59.0} \\

        \cmidrule{2-9}
        & POET-L        &  \ph{76.6} $|$ \ph{76.3} & \ph{45.6} $|$ \ph{44.6} & \ph{67.6} $|$ \ph{69.7} & \ph{44.4} $|$ \ph{41.7} & \ph{49.2} $|$ \ph{44.0} & \ph{51.2} & \ph{50.9} \\
        \rowcolor{highlightclr} \cellcolor{white} 
        \multirow{-7}{*}{\rotatebox[origin=c]{90}{\parbox[c]{2.4cm}{\centering Numerate LMs}}}
        & \ \ + \teabreac \icon &  \pg{79.1} $|$ \pg{78.6} & \pg{47.5} $|$ \pg{45.0} & \pg{72.2} $|$ \pg{71.5} & \pg{46.2} $|$ \pg{44.0} & \pg{51.9} $|$ \pg{48.6} & \pg{55.4} & \pg{54.8} \\

        \bottomrule
\end{tabular}
}

\caption{EM scores of in-distribution and robustness evaluation of language models (LMs) with and without \icon \xspace \teabreac pretraining on dev and test sets. Pretraining LMs on \teabreac improves their in-distribution performance and robustness across multiple QA datasets, for both plain and numerate LMs. In-distribution evaluation scores are (dev $|$ test) scores. Robustness evaluations are on test-only contrast sets. The suffixes `-3B', `-L' and `-S' refer to model sizes 3B, large and small, respectively. \pg{Green} (underlined) indicates \teabreac pretraining improves the underlying model's performance, while \pr{red} (not underlined) indicates it does not.}
\label{table:main-results-in-em}
\end{table*}

\section*{\teabreac Ablations on DROP dev set}
\label{sec:teabreac-ablations-drop-dev}

\begin{figure}[t]
    \centering
	\includegraphics[width=0.45\textwidth]{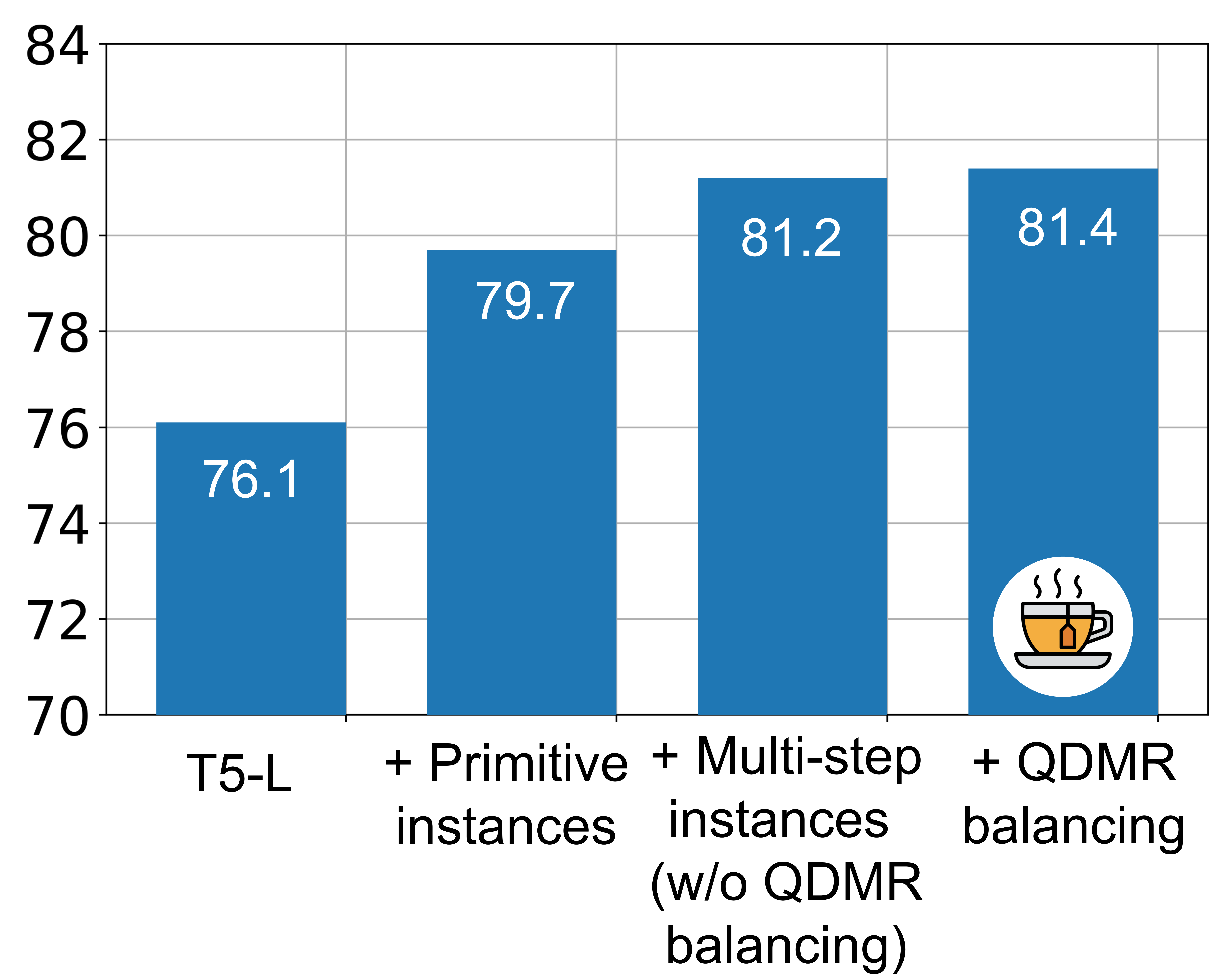}
	\caption{\teabreac Ablations: All the three aspects of \teabreac pretraining data contribute to overall performance: (i) primitive QA instances (ii) multi-step QA instances (iii) balancing of QDMRs to construct the multi-step QA dataset. The results are F1 scores on DROP dev set. The effect on DROP dev set is less prominent than in DROP CS and BPB sets, potentially due to shortcuts in DROP dev set.}
	\label{fig:teabreac-ablations-drop-dev}
\end{figure}

\teabreac ablation on DROP dev set is provided in Fig.~\ref{fig:teabreac-ablations-drop-dev}.

\section{Implementation Details}
\label{sec:implementation-details}

We train all models on a RTX A6000 (48GB) GPU. We pretrain on \teabreac by sampling a batch from multi-step and primitive synthetic instances in an alternating fashion. The hyperparameters for pretraining and fine-tuning are given in Table~\ref{tab:hyperparameters}. The only hyperparameter we sweeped over is learning rate ($10^{-5}$, $5 \times 10^{-5}$, $10^{-4}$, $5 \times 10^{-4}$, $10^{-3}$). The number of epochs were set to a large number with early stopping based on validation score. We've used Adafactor optimizer for all our experiments~\cite{adafactor}. We selected training hyper-parameter (learning rate) for each baseline model and dataset, based on the validation set performance. Our fine-tuning experiments using models pretrained on \teabreac use this identical learning rate.

\begin{table}[ht]
    \centering
    \setlength{\tabcolsep}{2.5pt}
    \begin{tabular}{cccccc}\toprule
        Model & Dataset & LR & Epochs & BS \\

        \midrule
        Bart-L        &  \teabreac & $10^{-5}$             & 20 & 16 \\
        T5-L          &  \teabreac & $10^{-4}$             & 20 & 8  \\
        T5-3B         &  \teabreac & $5\times10^{-5}$      & 10 & 8  \\
        NT5-S         &  \teabreac & $10^{-3}$             & 20 & 32 \\
        \preasm-L     &  \teabreac & $5\times10^{-5}$      & 20 & 8  \\
        \poet-L       &  \teabreac & $10^{-5}$             & 20 & 16 \\

        \midrule
        Bart-L        &  DROP & $10^{-5}$             & 20 & 8 \\
        T5-L          &  DROP & $10^{-4}$             & 20 & 8 \\
        T5-3B         &  DROP & $5\times10^{-5}$      & 10 & 8 \\
        NT5-S         &  DROP & $10^{-3}$             & 40 & 32 \\
        \preasm-L     &  DROP &  $5\times10^{-5}$     & 20 & 8 \\
        \poet-L       &  DROP &  $10^{-5}$            & 20 & 8 \\

        Bart-L        &  TAT-QA   & $10^{-5}$             & 20 & 8 \\
        T5-L          &  TAT-QA   & $10^{-4}$             & 20 & 8 \\
        T5-3B         &  TAT-QA   & $5\times10^{-5}$      & 10 & 8 \\
        NT5-S         &  TAT-QA   & $10^{-3}$             & 40 & 32 \\
        \preasm-L     &  TAT-QA   & $5\times10^{-5}$      & 20 & 8 \\
        \poet-L       &  TAT-QA   &  $10^{-5}$            & 20 & 8 \\

        Bart-L        &  IIRC     & $10^{-5}$         & 20 & 8 \\
        T5-L          &  IIRC     & $10^{-4}$         & 20 & 8 \\
        T5-3B         &  IIRC     & $5\times10^{-5}$  & 10 & 8 \\
        NT5-S         &  IIRC     & $10^{-3}$         & 40 & 32 \\
        \preasm-L     &  IIRC     & $5\times10^{-5}$  & 20 & 8 \\
        \poet-L       &  IIRC     &  $10^{-5}$        & 20 & 8 \\

        Bart-L        &  \numglue & $10^{-5}$         & 20 & 8 \\
        T5-L          &  \numglue & $10^{-4}$         & 20 & 8 \\
        T5-3B         &  \numglue & $5\times10^{-5}$  & 10 & 8 \\
        NT5-S         &  \numglue & $10^{-3}$         & 40 & 32 \\
        \preasm-L     &  \numglue & $5\times10^{-5}$  & 20 & 8 \\
        \poet-L       &  \numglue &  $10^{-5}$        & 20 & 8 \\

        \bottomrule
    \end{tabular}
    \caption{
    \textbf{Top:} Hyperparameters (HPs) for pretraining LMs on \teabreac. For large and 3B sized models, each epoch constitutes 200000/batch-size steps. For the small sized model (NT5-S), each epoch constitutes 2000000/batch-size steps. For each step, we uniformly randomly sample a batch of \teabreac multi-step instances or primitive instances. We've used identical HPs for pretraining ablations discussed in \S~\ref{subsec:teabreac-ablations}.
    \textbf{Bottom:} HPs for fine-tuning LMs on target datasets. We use the same HPs for fine-tuning LMs with or without \teabreac pretraining. The HPs for IIRC-G and IIRC-R experiments are the same. LR is learning rate and BS is batch size.}
    \label{tab:hyperparameters}
\end{table}

\section{Examples of Multi-Step QA Instances}
\label{sec:more-multihop-examples}
\begin{figure*}[ht]
    \centering
    \includegraphics[width=1\textwidth]{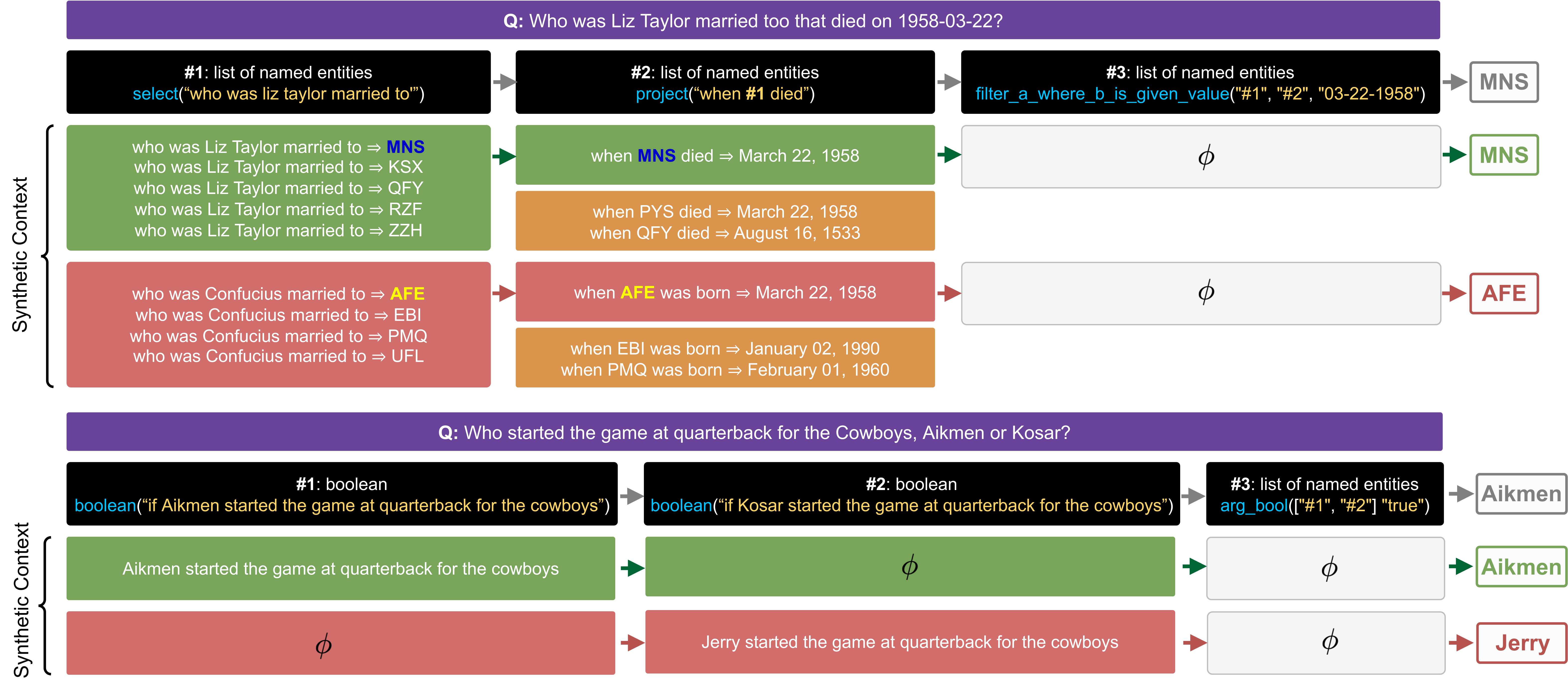}
    \caption{Synthetic reading comprehension QA instances involving \texttt{project} (top) and \texttt{boolean} (bottom) primitives.}
    \label{fig:synthetic-instance-examples-apdx}
\end{figure*}

Example multi-step QA instances with \texttt{project} and \texttt{boolean} primitives are shown in Fig.~\ref{fig:synthetic-instance-examples-apdx}.

\section{List of Primitives (Python Functions)}
\label{sec:primitives-list}
List of primitives (python functions) and a corresponding example is given in Table~\ref{tab:primitives}.

\fontsize{10}{10}\selectfont

\clearpage
\onecolumn
\renewcommand*{\arraystretch}{1.9}
\begin{longtable}{p{5.2cm}p{9.5cm}}
\caption{List of primitives (python functions) and a corresponding example.}\\

\toprule
\textbf{Primitive} & \textbf{Example} \\
\midrule
\endfirsthead

\multicolumn{2}{c}%
{\tablename\ \thetable\ -- \textit{Continued from previous page}} \\
\toprule
\textbf{Primitive} & \textbf{Example} \\
\midrule
\endhead

\hline \multicolumn{2}{r}{\textit{Continued on next page}} \\
\endfoot
\hline
\endlastfoot

        compare\_numbers &
        \begin{minipage}{8cm}
            compare\_numbers(\#1, \#2, ``>") $\Rightarrow$ False \\ \\
            \textbf{State}: \\
            \#1: 25 \\
            \#2: 28
        \end{minipage} \\

        \midrule
        compare\_dates &
        \begin{minipage}{8cm}
            compare\_dates(\#1, \#2, ``>") $\Rightarrow$ False \\ \\
            \textbf{State}: \\
            \#1: 25 Jan 2012 \\
            \#2: 28 Jan 2012
        \end{minipage} \\

        \midrule
        maximum\_date &
        \begin{minipage}{8cm}
            maximum\_date([\#1, \#2]) $\Rightarrow$ 28 Jan 2012 \\ \\
            \textbf{State}: \\
            \#1: 25 Jan 2012 \\
            \#2: 28 Jan 2012
        \end{minipage} \\

        \midrule
        minimum\_date &
        \begin{minipage}{8cm}
            minimum\_date([\#1, \#2]) $\Rightarrow$ 25 Jan 2012 \\ \\
            \textbf{State}: \\
            \#1: 25 Jan 2012 \\
            \#2: 28 Jan 2012
        \end{minipage} \\

        \midrule
        date\_subtraction &
        \begin{minipage}{8cm}
            date\_subtraction(\#1, \#2, ``days") $\Rightarrow$ 3 \\ \\
            \textbf{State}: \\
            \#1: 25 Jan 2012 \\
            \#2: 28 Jan 2012
        \end{minipage} \\

        \midrule
        arg\_maximum\_date &
        \begin{minipage}{8cm}
            arg\_maximum\_date([\#1, \#2]) $\Rightarrow$ \#2 \\ \\
            \textbf{State}: \\
            \#1: 25 Jan 2012 \\
            \#2: 28 Jan 2012
        \end{minipage} \\

        \midrule
        arg\_minimum\_date &
        \begin{minipage}{8cm}
            arg\_minimum\_date([\#1, \#2]) $\Rightarrow$ \#1 \\ \\
            \textbf{State}: \\
            \#1: 25 Jan 2012 \\
            \#2: 28 Jan 2012
        \end{minipage} \\

        \midrule
        arg\_bool &
        \begin{minipage}{8cm}
            arg\_bool([\#1, \#2], ``true") $\Rightarrow$ \#1 \\ \\
            \textbf{State}: \\
            \#1: True \\
            \#2: False
        \end{minipage} \\

        \midrule
        count &
        \begin{minipage}{8cm}
            count(\#1) $\Rightarrow$ 3 \\ \\
            \textbf{State}: \\
            \#1: [ABC, XZE, PQR]
        \end{minipage} \\

        \midrule
        addition &
        \begin{minipage}{8cm}
            addition(\#1) $\Rightarrow$ 2657.3 \\ \\
            \textbf{State}: \\
            \#1: [3, 2564.2, 90.1]
        \end{minipage} \\

        \midrule
        subtraction &
        \begin{minipage}{8cm}
            subtraction(100, \#1): 75 \\ \\
            \textbf{State}: \\
            \#1: 25
        \end{minipage} \\

        \midrule
        multiplication &
        \begin{minipage}{8cm}
            multiplication(\#1, 5): 125 \\ \\
            \textbf{State}: \\
            \#1: 25
        \end{minipage} \\

        \midrule
        division &
        \begin{minipage}{8cm}
            division(\#1, 100): 254.2 \\ \\
            \textbf{State}: \\
            \#1: 25420
        \end{minipage} \\

        \midrule
        mean &
        \begin{minipage}{8cm}
            mean(\#1) $\Rightarrow$ 885.8 \\ \\
            \textbf{State}: \\
            \#1: [3, 2564.2, 90.1]
        \end{minipage} \\

        \midrule
        maximum\_number &
        \begin{minipage}{8cm}
            maximum\_number(\#1) $\Rightarrow$ 2564.2 \\ \\
            \textbf{State}: \\
            \#1: [3, 2564.2, 90.1]
        \end{minipage} \\

        \midrule
        minimum\_number &
        \begin{minipage}{8cm}
            minimum\_number(\#1) $\Rightarrow$ 3 \\ \\
            \textbf{State}: \\
            \#1: [3, 2564.2, 90.1]
        \end{minipage} \\

        \midrule
        arg\_maximum\_number &
        \begin{minipage}{8cm}
            arg\_maximum\_number([\#1, \#2, \#3]) $\Rightarrow$ \#2 \\ \\
            \textbf{State}: \\
            \#1: 3 \\
            \#2: 2564.2 \\
            \#3: 90.1
        \end{minipage} \\

        \midrule
        arg\_minimum\_number &
        \begin{minipage}{8cm}
            arg\_minimum\_number([\#1, \#2, \#3]) $\Rightarrow$ \#1 \\ \\
            \textbf{State}: \\
            \#1: 3 \\
            \#2: 2564.2 \\
            \#3: 90.1
        \end{minipage} \\



        \midrule
        kth\_highest &
        \begin{minipage}{8cm}
            kth\_highest(\#1, 2) $\Rightarrow$ 90.1 \\ \\
            \textbf{State}: \\
            \#1: [3, 2564.2, 90.1]
        \end{minipage} \\

        \midrule
        kth\_lowest &
        \begin{minipage}{8cm}
            kth\_lowest(\#1, 2) $\Rightarrow$ 90.1 \\ \\
            \textbf{State}: \\
            \#1: [3, 2564.2, 90.1]
        \end{minipage} \\

        \midrule
        are\_items\_same &
        \begin{minipage}{8cm}
            are\_items\_same(\#1, \#2) $\Rightarrow$ False \\ \\
            \textbf{State}: \\
            \#1: ABC \\
            \#2: EDX
        \end{minipage} \\

        \midrule
        are\_items\_different &
        \begin{minipage}{8cm}
            are\_items\_different(\#1, \#2) $\Rightarrow$ True \\ \\
            \textbf{State}: \\
            \#1: ABC \\
            \#2: EDX
        \end{minipage} \\

        \midrule
        filter\_a\_where\_b\_is\_max\_num &
        \begin{minipage}{8cm}
            filter\_a\_where\_b\_is\_max\_num(\#1, \#2) $\Rightarrow$ PQR \\ \\
            \textbf{State}: \\
            \#1: [ABC, PQR, MNZ] \\
            \#2: [3, 2564.2, 90.1]
        \end{minipage} \\

        \midrule
        filter\_a\_where\_b\_is\_min\_num &
        \begin{minipage}{8cm}
            filter\_a\_where\_b\_is\_min\_num(\#1, \#2) $\Rightarrow$ ABC \\ \\
            \textbf{State}: \\
            \#1: [ABC, PQR, MNZ] \\
            \#2: [3, 2564.2, 90.1]
        \end{minipage} \\

        \midrule
        filter\_a\_where\_b\_is\_given\_value &
        \begin{minipage}{8cm}
            filter\_a\_where\_b\_is\_given\_value(\#1, \#2, MNO) $\Rightarrow$ ABC \\ \\
            \textbf{State}: \\
            \#1: [ABC, PQR, MNZ] \\
            \#2: [MNO, XER, OIY]
        \end{minipage} \\

        \midrule
        filter\_a\_where\_b\_is\_compared\_to &
        \begin{minipage}{8cm}
            filter\_a\_where\_b\_is\_compared\_to(\#1, \#2, 80, >) $\Rightarrow$ [PQR, MNZ] \\ \\
            \textbf{State}: \\
            \#1: [ABC, PQR, MNZ] \\
            \#2: [3, 2564.2, 90.1]
        \end{minipage} \\

        \midrule
        filter\_a\_where\_b\_is\_in\_range &
        \begin{minipage}{8cm}
            filter\_a\_where\_b\_is\_in\_range\_num(\#1, \#2, 80, 100) $\Rightarrow$ [MNZ] \\ \\
            \textbf{State}: \\
            \#1: [ABC, PQR, MNZ] \\
            \#2: [3, 2564.2, 90.1]
        \end{minipage} \\

        \midrule
        filter\_a\_where\_b\_is\_compared\_to\_date &
        \begin{minipage}{8cm}
            filter\_a\_where\_b\_is\_compared\_to\_date(\#1, \#2, 25 Feb 2012, >) $\Rightarrow$ [PQR, MNZ] \\ \\
            \textbf{State}: \\
            \#1: [ABC, PQR, MNZ] \\
            \#2: [25 Jan 2012, 18 March 2012, 13 Oct 2019]
        \end{minipage} \\

        \midrule
        filter\_a\_where\_b\_is\_in\_range\_date &
        \begin{minipage}{8cm}
            filter\_a\_where\_b\_is\_in\_range\_date(\#1, \#2, 25 Feb 2012, 1 Nov 2021, 100) $\Rightarrow$ [PQR, MNZ] \\ \\
            \textbf{State}: \\
            \#1: [ABC, PQR, MNZ] \\
            \#2: [25 Jan 2012, 18 March 2012, 13 Oct 2019]
        \end{minipage} \\

        \midrule
        filter\_a\_where\_b\_is\_max\_date &
        \begin{minipage}{8cm}
            filter\_a\_where\_b\_is\_max\_date(\#1, \#2) $\Rightarrow$ MNZ \\ \\
            \textbf{State}: \\
            \#1: [ABC, PQR, MNZ] \\
            \#2:  [25 Jan 2012, 18 March 2012, 13 Oct 2019]
        \end{minipage} \\

        \midrule
        filter\_a\_where\_b\_is\_min\_date &
        \begin{minipage}{8cm}
            filter\_a\_where\_b\_is\_min\_date(\#1, \#2) $\Rightarrow$ ABC \\ \\
            \textbf{State}: \\
            \#1: [ABC, PQR, MNZ] \\
            \#2:  [25 Jan 2012, 18 March 2012, 13 Oct 2019]
        \end{minipage} \\

        \midrule
        grouped\_count &
        \begin{minipage}{8cm}
            grouped\_count(\#1, \#2) $\Rightarrow$ {ABC: 2, XYI: 2, PQR: 1} \\ \\
            \textbf{State}: \\
            \#1: [ABC, XYI, ABC, PQR, XYI] \\
            \#2: [UIQ, QWA, OUE, UHI, RVC]
        \end{minipage} \\

        \midrule
        grouped\_sum &
        \begin{minipage}{8cm}
            grouped\_sum(\#1, \#2) $\Rightarrow$ {ABC: 4, XYI: 7, PQR: 4} \\ \\
            \textbf{State}: \\
            \#1: [ABC, XYI, ABC, PQR, XYI] \\
            \#2: [1, 2, 3, 4, 5]
        \end{minipage} \\

        \midrule
        grouped\_mean &
        \begin{minipage}{8cm}
            grouped\_mean(\#1, \#2) $\Rightarrow$ {ABC: 2, XYI: 3.5, PQR: 4} \\ \\
            \textbf{State}: \\
            \#1: [ABC, XYI, ABC, PQR, XYI] \\
            \#2: [1, 2, 3, 4, 5]
        \end{minipage} \\

        \midrule
        union &
        \begin{minipage}{8cm}
            union(\#1, \#2, \#3) $\Rightarrow$ [ABC, PQR, MNO, JHI, KMR] \\ \\
            \textbf{State}: \\
            \#1: [ABC, PQR] \\
            \#2: [MNO] \\
            \#3: [JHI, KMR]
        \end{minipage} \\

        \midrule
        intersection &
        \begin{minipage}{8cm}
            intersection(\#1, \#2) $\Rightarrow$ [PQR] \\ \\
            \textbf{State}: \\
            \#1: [ABC, PQR, MNO] \\
            \#2: [PQR]
        \end{minipage} \\

        \midrule
        arg\_intersection &
        \begin{minipage}{8cm}
            arg\_intersection(\#1, \#2, \#3) $\Rightarrow$ [WEC] \\ \\
            \textbf{State}: \\
            \#1: [XYI, ORE, WEC] \\
            \#2: [ABC, PQR, MNO] \\
            \#3: [null, null, MNO]
        \end{minipage} \\

        \midrule
        list\_subtraction &
        \begin{minipage}{8cm}
            list\_subtraction(\#1, \#2) $\Rightarrow$ [XYI, WEC] \\ \\
            \textbf{State}: \\
            \#1: [XYI, ORE, WEC] ;
            \#2: [ORE]
        \end{minipage} \\

        \midrule
        logical\_and &
        \begin{minipage}{8cm}
            logical\_and(\#1, \#2) $\Rightarrow$ False \\ \\
            \textbf{State}: \\
            \#1: False ;
            \#2: True
        \end{minipage} \\

        \midrule
        logical\_or &
        \begin{minipage}{8cm}
            logical\_or(\#1, \#2) $\Rightarrow$ True \\ \\
            \textbf{State}: \\
            \#1: False ;
            \#2: True
        \end{minipage} \\

        \midrule
        select &
        \begin{minipage}{8cm}
            select("touchdowns by Edwards") $\Rightarrow$ [ABC, DXE, FGH] \\ \\
            \textbf{Facts in context}: \\
            touchdowns by Edwards $\Rightarrow$ ABC \\
            touchdowns by Edwards $\Rightarrow$ DXE \\
            touchdowns by Edwards $\Rightarrow$ FGH
        \end{minipage} \\

        \midrule
        filter &
        \begin{minipage}{8cm}
            filter(``\#1 from 1st quarter") $\Rightarrow$ [ABC, DXE] \\ \\
            \textbf{State}: \\
            \#1: [ABC, DXE] \\ \\
            \textbf{Facts in context}: \\
            what is from 1st quarter? $\Rightarrow$ ABC \\
            what is from 1st quarter? $\Rightarrow$ DXE \\
            what is from 1st quarter? $\Rightarrow$ MNF \\
            what is from 1st quarter? $\Rightarrow$ IOU
        \end{minipage} \\

        \midrule
        project &
        \begin{minipage}{8cm}
            project(``when \#1 died") $\Rightarrow$ [March 22, 1958] \\ \\
            \textbf{State}: \\
            \#1: [MNS] \\ \\
            \textbf{Facts in context}: \\
            when PYS died $\Rightarrow$ March 22, 1958 \\
            when MNS died $\Rightarrow$ March 22, 1958 \\
            when QFY died $\Rightarrow$ August 16, 1533
        \end{minipage} \\

        \midrule
        boolean &
        \begin{minipage}{8cm}
            boolean(``if Aikmen started the game at quarterback for the cowboys") $\Rightarrow$ True \\ \\
            \textbf{Facts in context}: \\
            Aikmen started the game at quarterback for the cowboys
        \end{minipage} 

\label{tab:primitives}
\end{longtable}

\clearpage
\twocolumn

\section{Examples of Instances for Individual Primitives}
\label{sec:primitives-examples}

Examples of template based QA instances for teaching individual primitives are given in Table~\ref{tab:primitives-examples}.

\fontsize{10}{10}\selectfont

\clearpage
\onecolumn
\renewcommand*{\arraystretch}{1.9}
\begin{longtable}{p{5.2cm}p{9.5cm}}
\caption{Examples QA instances for individual primitives (python functions)}\\

\toprule
\textbf{Primitive} & \textbf{Example} \\
\midrule
\endfirsthead

\multicolumn{2}{c}%
{\tablename\ \thetable\ -- \textit{Continued from previous page}} \\
\toprule
\textbf{Primitive} & \textbf{Example} \\
\midrule
\endhead

\hline \multicolumn{2}{r}{\textit{Continued on next page}} \\
\endfoot
\hline
\endlastfoot

        compare\_numbers &
        \begin{minipage}{8cm}
\textbf{Quesion:} Is 984,486.24 greater than 594147.75? \\
\textbf{Context:}  \\
\textbf{Answer :} ['yes']
        \end{minipage} \\

        \midrule
        compare\_dates &
        \begin{minipage}{8cm}
\textbf{Quesion:} Is 1934-9-4 greater than 27 May 1899? \\
\textbf{Context:}  \\
\textbf{Answer :} ['yes']
        \end{minipage} \\

        \midrule
        maximum\_date &
        \begin{minipage}{8cm}
\textbf{Quesion:} Which of the following dates come later? \\
\textbf{Context:} 11/30/1690 , 1690-05-17 \\
\textbf{Answer :} ['November 30, 1690']
        \end{minipage} \\

        \midrule
        minimum\_date &
        \begin{minipage}{8cm}
\textbf{Quesion:} Which of the following dates come before the other? \\
\textbf{Context:} 1925-4-12 , 18 Apr 1696 \\
\textbf{Answer :} ['April 18, 1696']
        \end{minipage} \\

        \midrule
        date\_subtraction &
        \begin{minipage}{8cm}
\textbf{Quesion:} How many days passed between 1567-6-29 and May 28, 1567? \\
\textbf{Context:}  \\
\textbf{Answer :} ['32']
        \end{minipage} \\

        \midrule
        arg\_maximum\_date &
        \begin{minipage}{8cm}
\textbf{Quesion:} Which event has highest date: OUM or NKE? \\
\textbf{Context:} Event OUM has date 1977-3-13. Event NKE has date November, 5 2011. \\
\textbf{Answer :} ['NKE']
        \end{minipage} \\

        \midrule
        arg\_minimum\_date &
        \begin{minipage}{8cm}
\textbf{Quesion:} Which event happened earliest: KSX or KBO or JJT? \\
\textbf{Context:} Event KSX has date 11/9/1705. Event KBO has date 04 Jul, 1786. Event JJT has date 04/11/1729. \\
\textbf{Answer :} ['KSX']
        \end{minipage} \\


        \midrule
        count &
        \begin{minipage}{8cm}
\textbf{Quesion:} How many total entities the following list has? \\
\textbf{Context:} DMX NQX LFD RJN AMG \\
\textbf{Answer :} ['5']
        \end{minipage} \\

        \midrule
        addition &
        \begin{minipage}{8cm}
\textbf{Quesion:} Given the list of numbers, give their total sum. \\
\textbf{Context:} 977.98 ; 710 ; seven ; 4.72 \\
\textbf{Answer :} ['1699.7']
        \end{minipage} \\

        \midrule
        subtraction &
        \begin{minipage}{8cm}
\textbf{Quesion:} What is 721,251 - 32561? \\
\textbf{Context:}  \\
\textbf{Answer :} ['688690']
        \end{minipage} \\

        \midrule
        multiplication &
        \begin{minipage}{8cm}
\textbf{Quesion:} If you multiply forty-eight with 41, what do you get? \\
\textbf{Context:}  \\
\textbf{Answer :} ['1968']
        \end{minipage} \\

        \midrule
        division &
        \begin{minipage}{8cm}
\textbf{Quesion:} What is 47 divided by 6 in nearest integer? \\
\textbf{Context:}  \\
\textbf{Answer :} ['7']
        \end{minipage} \\

        \midrule
        mean &
        \begin{minipage}{8cm}
\textbf{Quesion:} What is the average of the following numbers in nearest integer? \\
\textbf{Context:} 172 ; 691 \\
\textbf{Answer :} ['431']
        \end{minipage} \\

        \midrule
        maximum\_number &
        \begin{minipage}{8cm}
\textbf{Quesion:} Given the following list, what is the largest number? \\
\textbf{Context:} 6603 ; 3.76 ; 636,337.65 ; 91.72 \\
\textbf{Answer :} ['636337.65']
        \end{minipage} \\

        \midrule
        minimum\_number &
        \begin{minipage}{8cm}
\textbf{Quesion:} What is the smallest of the following numbers? \\
\textbf{Context:} 60,810.74 ; 2.24 ; 48.8 \\
\textbf{Answer :} ['2.24']
        \end{minipage} \\

        \midrule
        arg\_maximum\_number &
        \begin{minipage}{8cm}
\textbf{Quesion:} Which entity has biggest value: ROJ or ZZH or KFI? \\
\textbf{Context:} Entity ROJ has value 91,889. Entity ZZH has value 0.93. Entity KFI has value 9,223.7. \\
\textbf{Answer :} ['ROJ']
        \end{minipage} \\

        \midrule
        arg\_minimum\_number &
        \begin{minipage}{8cm}
\textbf{Quesion:} Which entity has lowest value: TXM or KPG or JLD? \\
\textbf{Context:} Entity TXM has value 195.35. Entity KPG has value 861878. Entity JLD has value 41. \\
\textbf{Answer :} ['JLD']
        \end{minipage} \\



        \midrule
        kth\_highest &
        \begin{minipage}{8cm}
\textbf{Quesion:} Give the 2nd maximum value of \#17? \\
\textbf{Context:} \#17 has values 20787.56, 8265.18. \#9 has values January 25, 1787, January 27, 1787, January 08, 1787, January 18, 1787. \#3 has values February 14, 1994. \#18 has values 3.47, 4692.13, 735.31. \\
\textbf{Answer :} ['8265.18']
        \end{minipage} \\

        \midrule
        kth\_lowest &
        \begin{minipage}{8cm}
\textbf{Quesion:} Which is the 3rd lowest value of \#1? \\
\textbf{Context:} \#7 has values July 24, 1506, July 04, 1506, July 02, 1506, July 15, 1506. \#1 has values 2, 9, 23866. \#11 has values KFI, DXK, TFM. \\
\textbf{Answer :} ['23866']
        \end{minipage} \\

        \midrule
        are\_items\_same &
        \begin{minipage}{8cm}
\textbf{Quesion:} Are the following entities the same? \\
\textbf{Context:} Jan 07, 1696 and 01-7-1696. \\
\textbf{Answer :} ['yes']
        \end{minipage} \\

        \midrule
        are\_items\_different &
        \begin{minipage}{8cm}
\textbf{Quesion:} Are the following entities different? \\
\textbf{Context:} HUU and 09-29-1771. \\
\textbf{Answer :} ['yes']
        \end{minipage} \\

        \midrule
        filter\_a\_where\_b\_is\_max\_num &
        \begin{minipage}{8cm}
\textbf{Quesion:} What entity has biggest value? \\
\textbf{Context:} Entity OGQ has value 59. Entity HDU has value 94. Entity KLM has value 28,742. Entity LGV has value 713. Entity KGH has value 701. Entity DXK has value 373. \\
\textbf{Answer :} ['KLM']
        \end{minipage} \\

        \midrule
        filter\_a\_where\_b\_is\_min\_num &
        \begin{minipage}{8cm}
\textbf{Quesion:} Which entity has the minimum value? \\
\textbf{Context:} Entity FYO has value 266. Entity XHY has value 199052. Entity EQO has value 534. \\
\textbf{Answer :} ['FYO']
        \end{minipage} \\

        \midrule
        filter\_a\_where\_b\_is\_given\_value &
        \begin{minipage}{8cm}
\textbf{Quesion:} Which entities with value equal to 6.45? \\
\textbf{Context:} Entity KSX has value 6.45. Entity NLV has value 887.41. Entity OJP has value 603145.31. \\
\textbf{Answer :} ['KSX']
        \end{minipage} \\

        \midrule
        filter\_a\_where\_b\_is\_compared\_to &
        \begin{minipage}{8cm}
\textbf{Quesion:} Entities that have value larger than 948768.92? \\
\textbf{Context:} Entity AFE has value 871781. Entity RQX has value 989,517.24. \\
\textbf{Answer :} ['RQX']
        \end{minipage} \\


        \midrule
        filter\_a\_where\_b\_is\_compared\_to\_date &
        \begin{minipage}{8cm}
\textbf{Quesion:} List the entities with date below Jul 20 1646? \\
\textbf{Context:} Entity ZBK has date 9-12-1560. Entity AGU has date July 17 1953. \\
\textbf{Answer :} ['ZBK']
        \end{minipage} \\


        \midrule
        filter\_a\_where\_b\_is\_max\_date &
        \begin{minipage}{8cm}
\textbf{Quesion:} Which entity has latest date? \\
\textbf{Context:} Entity SML has value 11-28-1882. Entity PYS has value Nov 19 1882. \\
\textbf{Answer :} ['SML']
        \end{minipage} \\

        \midrule
        filter\_a\_where\_b\_is\_min\_date &
        \begin{minipage}{8cm}
\textbf{Quesion:} What entity has least recent date? \\
\textbf{Context:} Entity SDA has value 5 March, 1523. Entity HXJ has value 14 March 1523. Entity RZO has value 1-26-1523. Entity ZMH has value 23 Jul, 1523. \\
\textbf{Answer :} ['RZO']
        \end{minipage} \\

        \midrule
        grouped\_count &
        \begin{minipage}{8cm}
\textbf{Quesion:} How many times do each of EBC, HNQ occur in \#14? \\
\textbf{Context:} \#14 has HNQ, EBC, HNQ. \#3 has OZB, LNW, LYP, AGU, HVP, SDA. \#17 has ULN, ZZH, RZO \\
\textbf{Answer :} ['1', '2']
        \end{minipage} \\

        \midrule
        grouped\_sum &
        \begin{minipage}{8cm}
\textbf{Quesion:} What are the addition of values for each of QWU, JLD? \\
\textbf{Context:} QWU has value 179541.17. JLD has value 6,641.78. JLD has value 3.15. QWU has value 6,053.93. QWU has value 44,251.33. JLD has value 411.83. \\
\textbf{Answer :} ['229846.43', '7056.76']
        \end{minipage} \\

        \midrule
        grouped\_mean &
        \begin{minipage}{8cm}
\textbf{Quesion:} For each of TKR, NLV, what are the mean of values in integers? \\
\textbf{Context:} TKR has value 929. TKR has value 737. TKR has value ninety-five. NLV has value 928. \\
\textbf{Answer :} ['587', '928']
        \end{minipage} \\

        \midrule
        union &
        \begin{minipage}{8cm}
\textbf{Quesion:} Give answer union of \#20, \#12, \#13? \\
\textbf{Context:} \#20 has answer 29.77. \#12 has answer KBE. \#11 has answer June 10, 1701. \#13 has answer January 23, 1503. \\
\textbf{Answer :} ['29.77', 'KBE', 'January 23, 1503']
        \end{minipage} \\

        \midrule
        intersection &
        \begin{minipage}{8cm}
\textbf{Quesion:} List the entities that occur in both \#10 and \#7? \\
\textbf{Context:} \#1 has entities ICU, WAT. \#10 has entities WAT, ICU. \#7 has entities WAT, ICU. \\
\textbf{Answer :} ['ICU', 'WAT']
        \end{minipage} \\

        \midrule
        arg\_intersection &
        \begin{minipage}{8cm}
\textbf{Quesion:} List the entities contain values common in both \#9 and \#20? \\
\textbf{Context:} Entity KBE has value UJI for \#20. Entity KLM has value ARU for \#20. Entity KBE has no value for \#9. Entity KLM has value ARU for \#9. \\
\textbf{Answer :} ['KLM']

        \end{minipage} \\


        \midrule
        logical\_and &
        \begin{minipage}{8cm}
\textbf{Quesion:} What is logical AND of the given booleans? \\
\textbf{Context:} True False \\
\textbf{Answer :} ['no']
        \end{minipage} \\

        \midrule
        logical\_or &
        \begin{minipage}{8cm}
\textbf{Quesion:} What is logical OR of the given booleans? \\
\textbf{Context:} False False \\
\textbf{Answer :} ['no']
        \end{minipage} \\

\label{tab:primitives-examples}
\end{longtable}

\clearpage
\twocolumn

\end{document}